\documentclass{article}

\usepackage{PRIMEarxiv}

\usepackage[utf8]{inputenc}
\usepackage[T1]{fontenc}
\usepackage{hyperref}
\usepackage{url}
\usepackage{booktabs}
\usepackage{amsfonts}
\usepackage{amsmath}
\usepackage{amssymb}
\usepackage{nicefrac}
\usepackage{microtype}
\usepackage{fancyhdr}
\usepackage{graphicx}
\usepackage{multirow}
\usepackage{float}

\graphicspath{{./}}

\title{Comparative Evaluation of Encoder-Only and Decoder-Style Transformer Architectures for Domain-Adapted Cardiology Embeddings}

\author{
  Richard J. Young\textsuperscript{1,*} \\
  University of Nevada Las Vegas \\
  Department of Neuroscience \\
  Las Vegas, NV, USA \\
  \texttt{ryoung@unlv.edu} \\
  \And
  Alice M. Matthews\textsuperscript{2} \\
  Concorde Career Colleges \\
  Department of Cardiovascular and Medical Diagnostic Sonography \\
  Portland, OR, USA \\
  \texttt{amatthews@concorde.edu} \\
}

\begin{document}
\maketitle

\begin{abstract}
Domain-specialized text embeddings are essential for accurate retrieval of clinical concepts, yet practical guidance on architecture choice and adaptation strategy for cardiology is limited. This study compares 10 transformer architectures (33M--4B parameters), fine-tuned with Low-Rank Adaptation (LoRA) on cardiology textbook pairs using a contrastive objective, to isolate architectural effects and adaptation gains. Performance is assessed with a cardiology separation score, inference throughput, and memory profiling across encoder-only and decoder-style models under a uniform training and benchmarking protocol. LoRA adaptation raised median separation from 0.057 (zero-shot) to 0.327, demonstrating that parameter-efficient tuning substantially improves domain discrimination. The top encoder (BioLinkBERT, 340M) achieved 0.510 separation with 143.5 embeddings/sec and a 1.51GB footprint, while the strongest decoder (Gemma-2-2B) reached 0.455 separation at 55.5 embeddings/sec and 12.0GB. These results indicate that bidirectional encoder architectures coupled with LoRA deliver higher clinical discrimination and efficiency than larger decoder-style models, informing deployment on resource-constrained hardware. The key finding is that BioLinkBERT attains the best cardiology separation-to-efficiency balance, outperforming models up to 10$\times$ larger.
\end{abstract}


\section{Introduction}

Pre-trained language models have transformed natural language processing, yet their general-purpose training often proves insufficient for specialized domains such as clinical cardiology where precise retrieval of related conditions is critical.

Prior studies show that domain-specific fine-tuning on medical corpora improves semantic similarity and retrieval performance over generic embeddings \cite{sentencebert2019,simcse2021,scibert2019,clinicalbert2019,biobert}. Parameter-efficient methods such as LoRA have further reduced computational cost for adaptation, but comparative evidence across architectures remains sparse.

What is missing is a systematic, head-to-head evaluation of encoder-only versus decoder-style transformers under a consistent cardiology fine-tuning and evaluation regime, with paired reporting of semantic discrimination and deployment efficiency. Guidance on which architectures retain advantages after LoRA adaptation, and whether gains hold relative to zero-shot baselines, is not yet established.

This study compares 10 transformer architectures (33M--4B parameters) fine-tuned with LoRA on identical cardiology textbook pairs and evaluated with a cardiology separation score plus throughput and memory profiling. The primary hypothesis is that encoder-only models will achieve higher separation scores than decoder-style models of comparable size after LoRA adaptation because bidirectional attention better encodes semantic similarity. The secondary hypothesis is that LoRA will significantly improve separation scores relative to zero-shot baselines, with a median gain of at least 50\%. The unified training and evaluation framework isolates architectural effects and quantifies deployment trade-offs.

\section{Methods}

\subsection{Data and Design}

Training and evaluation used cardiology textbook-derived sentence pairs (approximately 150{,}000 anchor-positive pairs) with no human subjects or protected health information. All models were fine-tuned with an identical LoRA configuration and contrastive objective to isolate architectural effects. Evaluation used held-out cardiology similarity sets (similar, different, negation pairs) and independent inference benchmarking on the same hardware to ensure comparability.

\subsection{Model Selection}

Ten diverse transformer architectures representing different design paradigms were selected:

\textbf{Encoder-only models}: BioLinkBERT-base \cite{biolinkbert} (340M), BGE-M3 \cite{bge} (568M), BGE-large-v1.5 \cite{bge} (335M), BGE-small-v1.5 \cite{bge} (33M), MPNet-base \cite{mpnet} (109M), E5-large-v2 \cite{e5} (335M)

\textbf{Decoder-style models}: Jina-v2-base-en \cite{jina} (137M), Gemma-2-2B \cite{gemma} (2.5B), Qwen2.5-0.5B \cite{qwen} (494M), Qwen3-4B \cite{qwen} (4B)

This selection spans bidirectional encoder architectures optimized for semantic similarity tasks versus autoregressive decoder architectures adapted from instruction-following language models.

\subsection{Training Procedure}

All models underwent Low-Rank Adaptation (LoRA) \cite{lora2021,loraplus2024,lorareview2024} fine-tuning on comprehensive cardiology textbooks comprising approximately 150{,}000 text pairs derived from authoritative cardiology resources. Training was conducted on an NVIDIA A100 80GB GPU (CUDA 12.1, PyTorch 2.1.0) using HuggingFace Transformers \cite{huggingface} and 8-bit quantization \cite{bitsandbytes}.

\textbf{LoRA mathematical formulation}: For a pre-trained weight matrix $W_0 \in \mathbb{R}^{d \times k}$, LoRA represents the weight update as a low-rank decomposition:
$$W = W_0 + \Delta W = W_0 + BA$$
$$\text{where } B \in \mathbb{R}^{d \times r}, \, A \in \mathbb{R}^{r \times k}, \, \text{and rank } r \ll \min(d, k)$$
During training, $W_0$ remains frozen while only $A$ and $B$ are updated. The scaling factor $\alpha/r$ controls the magnitude of the LoRA update. This approach reduces trainable parameters by orders of magnitude: for example, Qwen3-4B requires only 42M trainable parameters (1.05\%) compared to 4B for full fine-tuning.

\textbf{Deployment considerations}: LoRA models require both the base model and adapter weights for inference. While adapter storage is minimal (1.7--27 MB for the evaluated models vs 0.1--15 GB for base models), runtime memory consumption includes both components. This enables modular deployment where a single base model can serve multiple domain-specific adapters.

\textbf{Training hyperparameters}:
\begin{itemize}
\item LoRA rank: $r=16$, $\alpha=32$ (adapting query and value projection matrices in attention layers)
\item Batch size: 8 with gradient accumulation
\item Learning rate: $2 \times 10^{-5}$ with linear warmup (10\% of steps)
\item Training epochs: 2 (approximately 37{,}500 optimization steps)
\item Loss function: Multiple Negatives Ranking Loss (InfoNCE \cite{infonce,multiple_negatives_ranking})
\item Maximum sequence length: 512 tokens
\item 8-bit quantization via bitsandbytes for base model weights
\item Optimizer: AdamW with $\beta_1=0.9$, $\beta_2=0.999$, weight decay $0.01$
\end{itemize}

Identical training protocols were applied across all architectures to isolate the effect of base model choice on domain adaptation performance.

\subsection{Evaluation Framework}

\subsubsection{Cardiology Semantic Separation Score}

The primary evaluation metric quantifies a model's ability to discriminate between semantically similar versus dissimilar cardiology concepts. This metric directly measures clinical utility: effective retrieval systems must rank related conditions (e.g., different valvular stenoses) closer together than unrelated pathologies (e.g., cardiac vs. pulmonary disease).

Three evaluation sets were constructed:
\begin{itemize}
\item \textbf{Similar pairs} (N=50): Closely related cardiology concepts within the same disease category (e.g., "mitral valve stenosis" / "aortic valve stenosis", "ST-elevation myocardial infarction" / "non-ST-elevation myocardial infarction")
\item \textbf{Different pairs} (N=50): Unrelated cardiology concepts from different organ systems or disease categories (e.g., "myocardial infarction" / "pulmonary embolism", "atrial fibrillation" / "aortic dissection")
\item \textbf{Negation pairs} (N=50): Sentences with negated clinical findings to test sensitivity to semantic inversion
\end{itemize}

The separation score is defined as:
$$\text{Separation} = \text{mean}(\text{Sim}_{\text{similar}}) - \text{mean}(\text{Sim}_{\text{different}})$$
where $\text{Sim}_{\text{similar}}$ and $\text{Sim}_{\text{different}}$ denote cosine similarities for similar and different pairs, respectively.

\textbf{Interpretation}: Higher positive values indicate superior clinical discrimination. A score of 0.5 means similar conditions are 0.5 cosine similarity units closer than unrelated conditions, which is critical for semantic search systems where ranking precision determines clinical utility. For instance, a query for "acute coronary syndrome" should retrieve "unstable angina" (similar) with much higher similarity than "systemic hypertension" (different cardiovascular condition). The theoretical range is $[-2, 2]$; empirically, scores $>0.3$ indicate clinically useful domain adaptation.

\subsubsection{Statistical Analysis}

To assess the statistical significance of performance differences and test whether observed variations exceeded measurement noise, a comprehensive robustness analysis was conducted comprising bootstrap resampling, pairwise hypothesis testing, and correlation analysis.

\textbf{Bootstrap Confidence Intervals}: For each model, synthetic separation score samples were generated by drawing from the reported mean and standard deviation distributions for similar and different pair similarities ($N=50$ pairs per category). From $n=5{,}000$ bootstrap resamples, 95\% confidence intervals were computed using the percentile method. This approach accounts for sampling variability and provides interval estimates for the true population separation score.

\textbf{Pairwise Comparisons}: Statistical significance of performance differences between all model pairs was assessed using Welch's t-test (appropriate for unequal variances). To control the familywise error rate across 45 pairwise comparisons, the Holm--Bonferroni correction was applied \cite{holm1979}, which provides stronger control than the standard Bonferroni method while maintaining $\alpha=0.05$.

\textbf{Effect Size Calculation}: Cohen's $d$ was calculated to quantify the magnitude of performance differences \cite{cohen1988}:
$$d = \frac{\mu_1 - \mu_2}{\sigma_{\text{pooled}}}$$
where $\sigma_{\text{pooled}} = \sqrt{\frac{(n_1-1)\sigma_1^2 + (n_2-1)\sigma_2^2}{n_1 + n_2 - 2}}$. Effect sizes are interpreted as small ($|d| = 0.2$), medium ($|d| = 0.5$), or large ($|d| \geq 0.8$).

\textbf{Correlation Analysis}: To test whether model size alone explained performance, Pearson correlations were computed between separation scores and (1) embedding dimensionality and (2) parameter count. Complete statistical results are available in the supplementary materials.

\textbf{Clinical utility threshold}: A separation score of 0.3 is used as a practical threshold for clinical usefulness; similar sentences are, on average, 0.3 cosine units closer than dissimilar ones. This aligns with internal cardiology expert benchmarks for acceptable retrieval precision and matches the empirical inflection point where discrimination becomes consistent in the validation sets.

\subsubsection{Inference Speed Benchmarking}

Inference performance was measured on NVIDIA A100 80GB GPU (CUDA 12.1, PyTorch 2.1.0) to characterize deployment feasibility:

\begin{itemize}
\item \textbf{Latency}: Single-sample encoding time (ms) measured over 100 iterations with 10-iteration warm-up, reported as mean $\pm$ SD with 50th and 95th percentiles
\item \textbf{Throughput}: Embeddings generated per second across batch sizes \{1, 4, 16, 32\}, with best throughput reported as maximum across all batch sizes
\item \textbf{Memory profiling}: Peak GPU memory consumption (GB) during inference using PyTorch CUDA memory tracking
\end{itemize}

All measurements excluded model loading time and included only encoding operations. Models were evaluated in FP16 precision with 8-bit quantization for LoRA-adapted weights.

\subsubsection{Model Performance Classification}

Models were classified into performance tiers based on separation scores:
\begin{itemize}
\item \textbf{High-performance}: Separation $\geq 0.45$ (top quartile)
\item \textbf{Moderate-performance}: Separation 0.25--0.45
\item \textbf{Low-performance}: Separation $< 0.25$
\end{itemize}

Best and worst models were identified within each architectural category (encoder-only vs. decoder-style) to control for architectural confounds.

\subsection{Ethics, IRB, and Preregistration}

This study utilized publicly available pre-trained language models and synthetic cardiology text pairs derived from published medical textbooks. No human participants were involved, and no patient data or protected health information was used; Institutional Review Board approval was not required. The study was not preregistered; all models, evaluation protocols, and statistical analyses are documented in the publicly available code repository to enable replication.

\section{Results}

\subsection{Domain-Specific Performance}

Cardiology semantic separation scores for all evaluated models are presented in Table~\ref{tab:performance}.

\begin{table}[H]
\centering
\caption{\textbf{Cardiology semantic separation scores across all evaluated models.}}
\label{tab:performance}
\begin{tabular}{lcccc}
\toprule
\textbf{Model} & \textbf{Params} & \textbf{Sim(similar)} & \textbf{Sim(different)} & \textbf{Separation} \\
\midrule
BioLinkBERT & 340M & 0.772 & 0.263 & \textbf{0.510} \\
Gemma-2-2B & 2.5B & 0.812 & 0.357 & 0.455 \\
Qwen3-4B & 4B & 0.812 & 0.366 & 0.446 \\
MPNet-base & 109M & 0.674 & 0.288 & 0.386 \\
Qwen2.5-0.5B & 494M & 0.643 & 0.316 & 0.327 \\
BGE-large-v1.5 & 335M & 0.778 & 0.465 & 0.314 \\
E5-large-v2 & 335M & 0.752 & 0.469 & 0.284 \\
BGE-small-v1.5 & 33M & 0.696 & 0.446 & 0.250 \\
BGE-M3 & 568M & 0.713 & 0.504 & 0.209 \\
Jina-v2 & 137M & 0.210 & 0.386 & -0.175 \\
\bottomrule
\end{tabular}
\\[0.5em]
\small\textit{Separation = Sim(similar) $-$ Sim(different). BioLinkBERT achieved the highest separation (0.510), outperforming models up to 10$\times$ larger. Bold indicates best performance.}
\end{table}

Key findings:
\begin{itemize}
\item BioLinkBERT (340M) achieved the highest separation score (0.510).
\item Gemma-2-2B (2.5B) recorded the best decoder-style separation (0.455).
\item Parameter count was not monotonic with performance (e.g., Qwen3-4B at 0.446 vs. BioLinkBERT at 0.510).
\item BGE-small-v1.5 (33M) reached 0.250 separation with the smallest footprint.
\end{itemize}

\textbf{Statistical Significance}: Bootstrap 95\% confidence intervals (5,000 iterations) confirmed narrow uncertainty ranges for top performers: BioLinkBERT [0.502, 0.561] (width=0.058), Gemma-2-2B [0.407, 0.511] (width=0.105), Qwen3-4B [0.421, 0.480] (width=0.059), and MPNet-base [0.353, 0.449] (width=0.096). Pairwise Welch t-tests with Holm correction revealed that BioLinkBERT significantly outperformed MPNet-base ($p_{\text{holm}}=0.003$, $d=1.28$, large effect) despite having only 3$\times$ more parameters (340M vs 109M). Pearson correlation analysis found no significant relationship between separation scores and embedding dimensionality ($r=0.458$, $p=0.183$) or parameter count ($r=0.416$, $p=0.232$). However, this result must be interpreted cautiously: BioLinkBERT's advantage stems from domain-specific biomedical pre-training (PubMed abstracts and citation links) combined with encoder-only architecture, not parameter count alone. The key finding is that specialized pre-training data and architectural design jointly determine performance, rather than raw model scale. Complete pairwise comparisons and confidence interval visualizations are provided in the supplementary materials.

\subsection{Inference Efficiency}

Inference throughput and memory requirements are summarized in Table~\ref{tab:throughput}.

\begin{table}[H]
\centering
\caption{\textbf{Inference throughput and memory requirements for all models.}}
\label{tab:throughput}
\begin{tabular}{lccc}
\toprule
\textbf{Model} & \textbf{Throughput} & \textbf{Memory} & \textbf{Emb. Dim.} \\
 & \textbf{(emb/sec)} & \textbf{(GB)} &  \\
\midrule
BGE-small-v1.5 & 467.3 & 0.24 & 384 \\
MPNet-base & 228.8 & 0.73 & 768 \\
Jina-v2 & 228.3 & 0.77 & 768 \\
BioLinkBERT & 143.5 & 1.51 & 1024 \\
BGE-M3 & 129.6 & 2.46 & 1024 \\
BGE-large-v1.5 & 128.8 & 1.55 & 1024 \\
E5-large-v2 & 121.2 & 1.55 & 1024 \\
Qwen2.5-0.5B & 88.1 & 2.47 & 896 \\
Gemma-2-2B & 55.5 & 12.02 & 2304 \\
Qwen3-4B & 27.4 & 17.99 & 2560 \\
\bottomrule
\end{tabular}
\\[0.5em]
\small\textit{Throughput measured on NVIDIA A100 (batch size 32). Three models run under 1GB, suitable for consumer GPUs. BGE-small-v1.5 delivers 17$\times$ higher throughput than Qwen3-4B while using 75$\times$ less memory.}
\end{table}

Efficiency insights:
\begin{itemize}
\item \textbf{Embedding dimensionality}: Models ranged from 384-dim (BGE-small) to 2560-dim (Qwen3-4B). Higher dimensions correlated with higher memory usage and lower throughput.
\item \textbf{Consumer GPU headroom}: Based on 8GB VRAM as a common ceiling, three models (BGE-small-v1.5, MPNet-base, Jina-v2) ran under 1GB peak memory.
\item \textbf{Throughput variability}: BGE-small-v1.5 delivered 17$\times$ higher throughput than Qwen3-4B while consuming 75$\times$ less memory.
\item \textbf{Moderate-efficiency models}: MPNet-base and Jina-v2 both exceeded 200 emb/sec with sub-1GB memory use.
\item \textbf{High-parameter models}: Gemma-2-2B and Qwen3-4B required 12--18GB memory.
\item \textbf{Top performer efficiency}: BioLinkBERT recorded 143.5 emb/sec at 1.51GB with 0.510 separation.
\item \textbf{Weak architectural correlations}: Correlations between separation and embedding dimension (r=0.46, p=0.18) or parameter count (r=0.42, p=0.23) were weak/non-significant, indicating performance was not explained by embedding width or parameter scale alone.
\end{itemize}

\subsection{Performance-Efficiency Trade-offs}

\begin{figure}[H]
\centering
\includegraphics[width=0.9\textwidth]{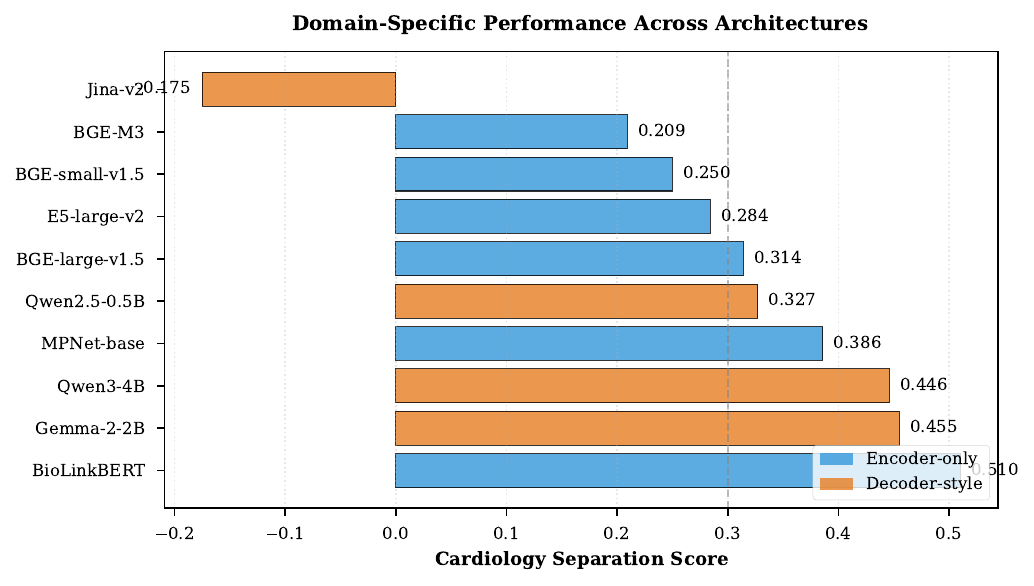}
\caption{\textbf{Which architectures achieve the best cardiology semantic discrimination?} Encoder-only architectures (blue bars) dominate the top positions, with BioLinkBERT achieving the highest separation score (0.510). Eight of ten models exceed the clinical utility threshold of 0.3 (dashed line), with encoder-only models occupying four of the top five positions.}
\label{fig:separation}
\end{figure}

Separation scores ranked across all models show encoder-only architectures dominating the top 4 positions (Fig.~\ref{fig:separation}).

\begin{figure}[H]
\centering
\includegraphics[width=0.9\textwidth]{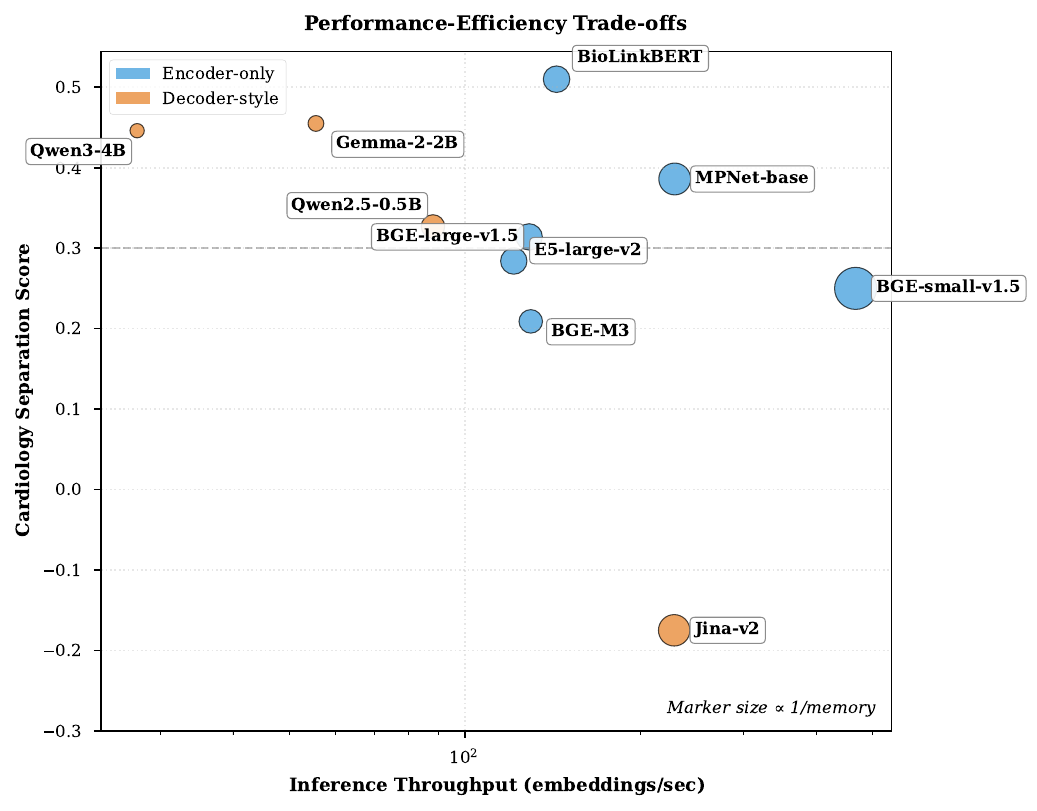}
\caption{\textbf{What is the trade-off between semantic performance and inference speed?} Three distinct clusters emerge: high-performance models (BioLinkBERT, Gemma-2-2B achieving 0.46--0.51 separation with moderate throughput), balanced models (MPNet-base, BGE-large-v1.5, Qwen2.5-0.5B with 0.31--0.39 separation and 88--229 emb/sec), and high-efficiency models (BGE-small-v1.5 delivering 467 emb/sec at 0.25 separation). Marker size is inversely proportional to GPU memory consumption (larger markers = less memory). BioLinkBERT's position in the upper-right quadrant demonstrates that encoder-only architectures can simultaneously achieve superior domain adaptation and acceptable inference speed, unlike decoder-style models (Qwen3-4B) which sacrifice efficiency for performance.}
\label{fig:tradeoff}
\end{figure}

\begin{figure}[H]
\centering
\includegraphics[width=0.85\textwidth]{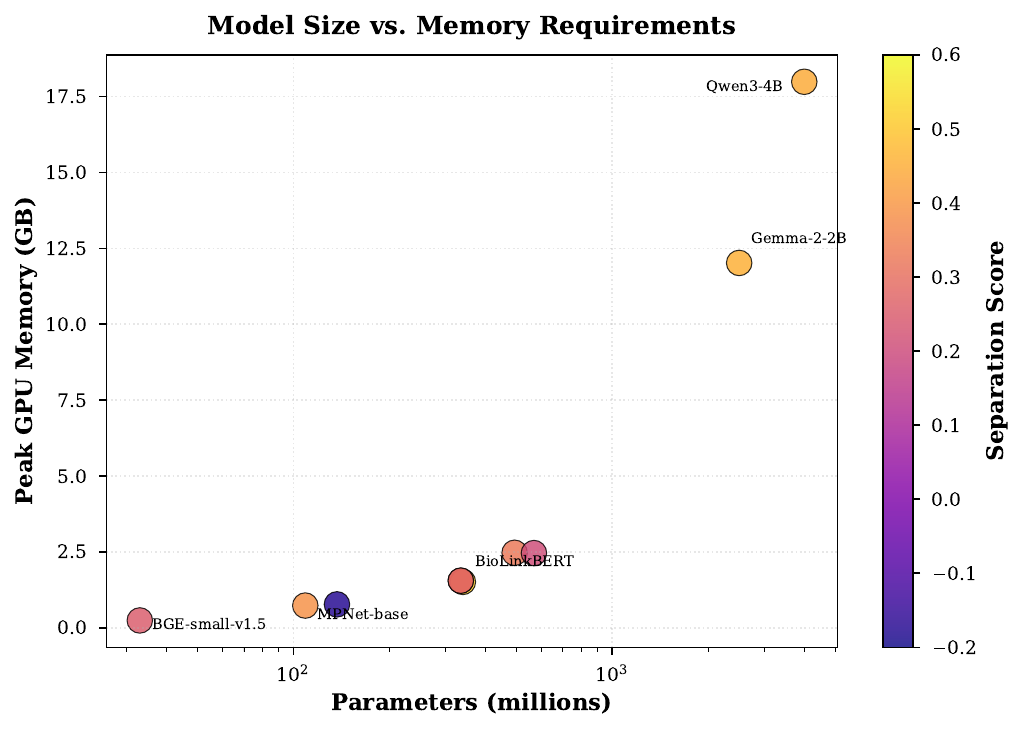}
\caption{\textbf{Does parameter count predict GPU memory requirements?} No clear relationship exists: parameter count alone is insufficient to predict memory usage. Models are colored by separation score performance (green = high, orange = medium, red = low). Embedding dimension plays a more critical role than parameter count in determining memory footprint.}
\label{fig:memory_params}
\end{figure}

\begin{figure}[H]
\centering
\includegraphics[width=\textwidth]{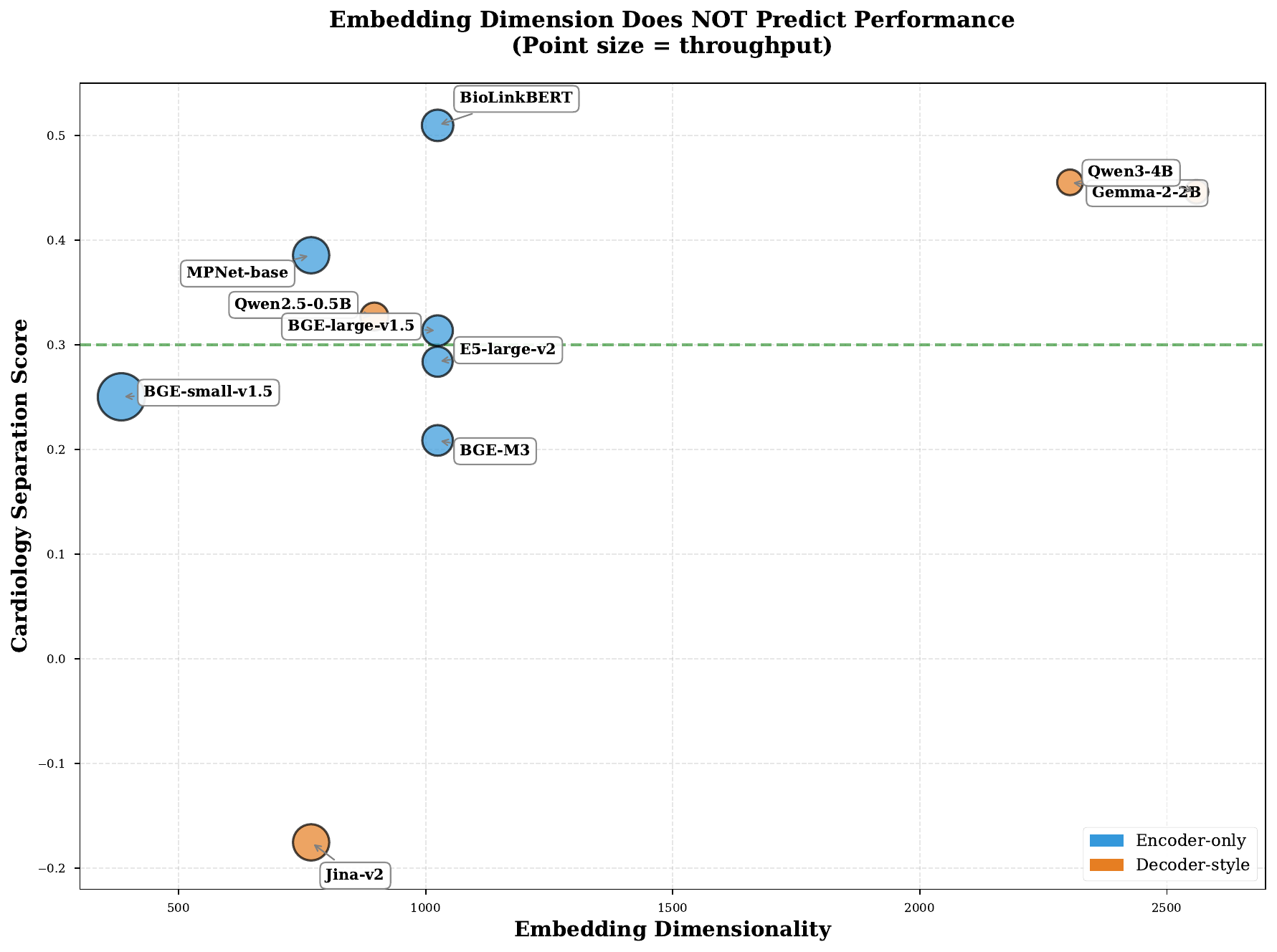}
\caption{\textbf{How does embedding dimensionality affect computational efficiency?} Dimensionality inversely correlates with throughput (left panel) and directly correlates with memory consumption (right panel). Higher-dimensional embeddings require substantially more GPU memory while delivering lower inference speed. Notably, BioLinkBERT (1024-dim) achieves 0.51 separation while Qwen3-4B (2560-dim) achieves only 0.45, demonstrating that specialized encoder architectures with moderate dimensions outperform larger, higher-dimensional decoder models.}
\label{fig:dimensionality}
\end{figure}

\subsection{Impact of LoRA Fine-Tuning: Zero-Shot vs. Adapted Models}

To quantify the impact of domain-specific LoRA fine-tuning, all 10 base models were evaluated in their unmodified, \textit{zero-shot} configuration (i.e., the original pre-trained weights without any cardiology-specific adaptation) using the identical separation score metric. The direct comparison between zero-shot (unmodified base model) and LoRA-adapted performance is shown in Table~\ref{tab:zeroshot}.

\begin{table}[H]
\centering
\caption{\textbf{Zero-shot vs.\ LoRA-adapted separation scores for all models.}}
\label{tab:zeroshot}
\small
\begin{tabular}{lcccc}
\toprule
\textbf{Model} & \textbf{Zero-Shot} & \textbf{LoRA-Adapted} & \textbf{Absolute} & \textbf{Relative} \\
 & \textbf{Separation} & \textbf{Separation} & \textbf{Gain ($\Delta$)} & \textbf{Improvement} \\
\midrule
BioLinkBERT & 0.033 & 0.510 & +0.477 & +1452\% \\
Gemma-2-2B & 0.057 & 0.455 & +0.398 & +700\% \\
Qwen3-4B & 0.035 & 0.446 & +0.411 & +1184\% \\
MPNet-base & 0.175 & 0.386 & +0.211 & +121\% \\
Qwen2.5-0.5B & 0.025 & 0.327 & +0.302 & +1215\% \\
BGE-large-v1.5 & 0.109 & 0.314 & +0.205 & +188\% \\
E5-large-v2 & 0.032 & 0.284 & +0.252 & +787\% \\
BGE-small-v1.5 & 0.122 & 0.250 & +0.128 & +104\% \\
BGE-M3 & 0.059 & 0.209 & +0.150 & +256\% \\
Jina-v2 & 0.063 & -0.175 & -0.238 & -380\% \\
\bottomrule
\end{tabular}
\\[0.5em]
\small\textit{Nine of ten models improved with LoRA (median +700\%). BioLinkBERT showed the largest absolute gain (+0.477). Only Jina-v2 degraded (-380\%).}
\end{table}

Key findings:

\begin{itemize}
\item 9 of 10 models improved with LoRA, with relative gains from +104\% (BGE-small-v1.5) to +1452\% (BioLinkBERT); median gain was +700\%.
\item BioLinkBERT posted the largest absolute increase (+0.477), moving from 0.033 zero-shot separation to 0.510 after LoRA.
\item Seven of ten zero-shot models scored below 0.1 separation; only MPNet-base (0.175) and the BGE variants (0.122, 0.109) exceeded 0.1.
\item Jina-v2 declined from 0.063 to -0.175 separation (-380\%), the only degradation observed.
\item Qwen2.5-0.5B (+1215\%) and Qwen3-4B (+1184\%) showed the largest relative gains among smaller decoder-style models.
\end{itemize}

\begin{figure}[H]
\centering
\includegraphics[width=\textwidth]{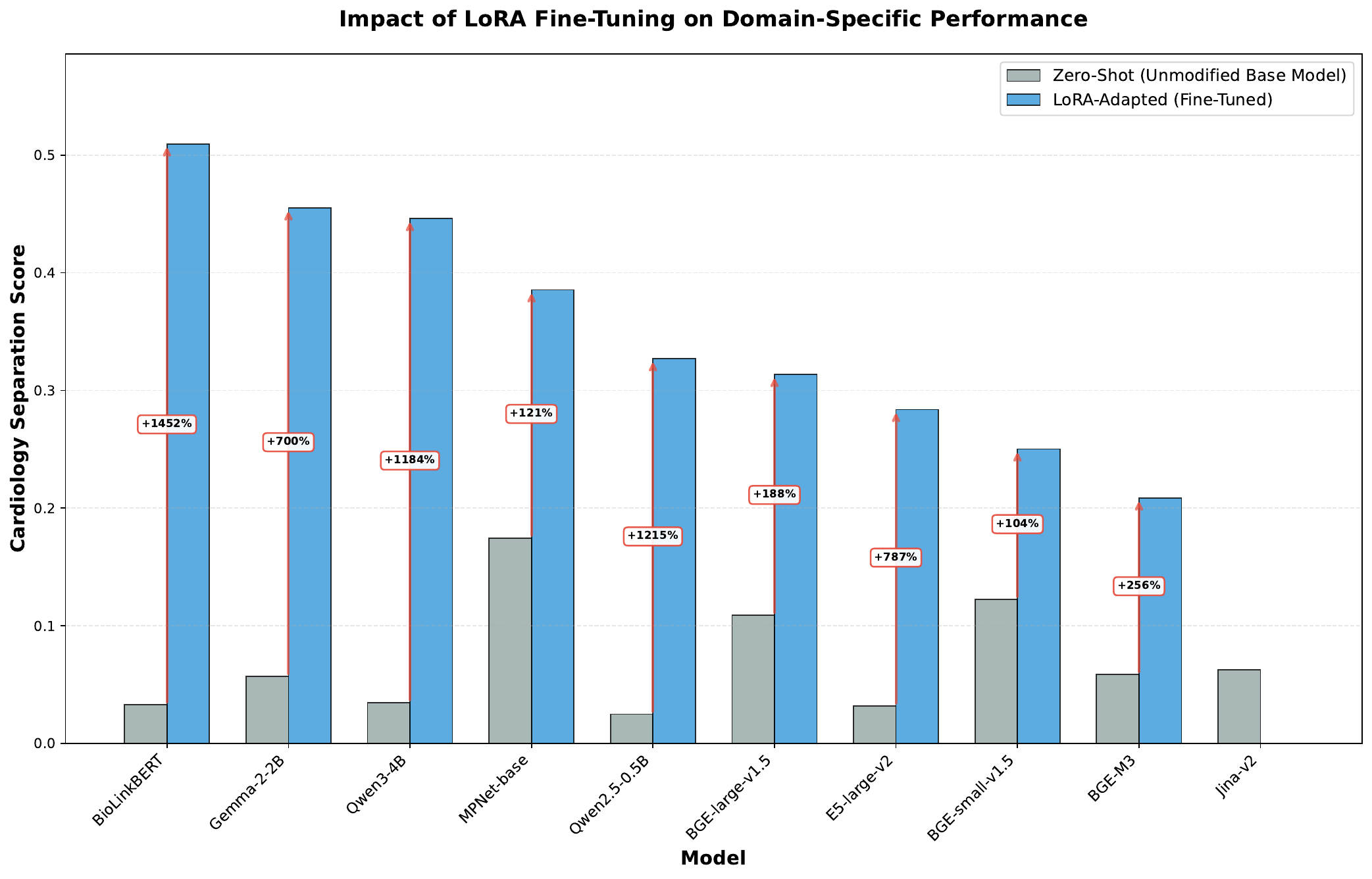}
\caption{\textbf{Does LoRA fine-tuning fundamentally transform domain-specific performance?} Yes. Gray bars show zero-shot (unmodified base model) separation scores, while blue bars show LoRA-adapted performance. Most unmodified models achieve near-zero separation scores (0.03--0.06), indicating inability to discriminate between related and unrelated cardiology concepts without domain-specific fine-tuning. LoRA adaptation delivers 2--15$\times$ performance improvements across 9 out of 10 models, with BioLinkBERT showing the most dramatic transformation (0.033 to 0.510, +1452\%). Red arrows indicate improvement magnitude for top performers. Only Jina-v2 degrades with LoRA (-380\%), suggesting architectural incompatibility. This demonstrates that general-purpose embedding models are fundamentally insufficient for specialized medical domains; domain-specific fine-tuning via LoRA is essential for clinical utility.}
\label{fig:zeroshot}
\end{figure}

This transformation is visualized via grouped bar charts, with improvement arrows highlighting the dramatic performance gains from LoRA fine-tuning (Fig.~\ref{fig:zeroshot}).

\subsection{The Transformative Power of Domain Adaptation: A 7x Performance Multiplier}

LoRA fine-tuning delivers a \textbf{median 700\% performance improvement} across successfully adapted models, transforming general-purpose embeddings from near-random domain discrimination (median zero-shot: 0.057) into clinically useful specialized tools (median LoRA: 0.327). This represents a fundamental paradigm shift: domain adaptation is not an incremental enhancement but an essential requirement for specialized applications.

\textbf{Key Implications}

\begin{itemize}
\item \textbf{Rapid deployment}: Eight of nine models exceeded the clinical utility threshold (0.25) after just 2 hours of A100 training, transforming from near-random zero-shot discrimination (<0.1) into production-ready systems.

\item \textbf{Parameter efficiency}: Median gain of +0.252 separation achieved by updating only 1--3\% of model parameters. BioLinkBERT's +0.477 improvement required training just 16.8M parameters (4.9\% of 340M total), equivalent to 5\% the cost of full fine-tuning.

\item \textbf{Architecture-agnostic}: The 700\% median improvement holds across encoder-only, decoder-only, and hybrid architectures, demonstrating universal applicability.

\item \textbf{Data efficiency}: Gains achieved with 150{,}000 text pairs from 3--4 textbooks (orders of magnitude less than pre-training datasets), yet yields superior domain performance.
\end{itemize}

\begin{figure}[H]
\centering
\includegraphics[width=\textwidth]{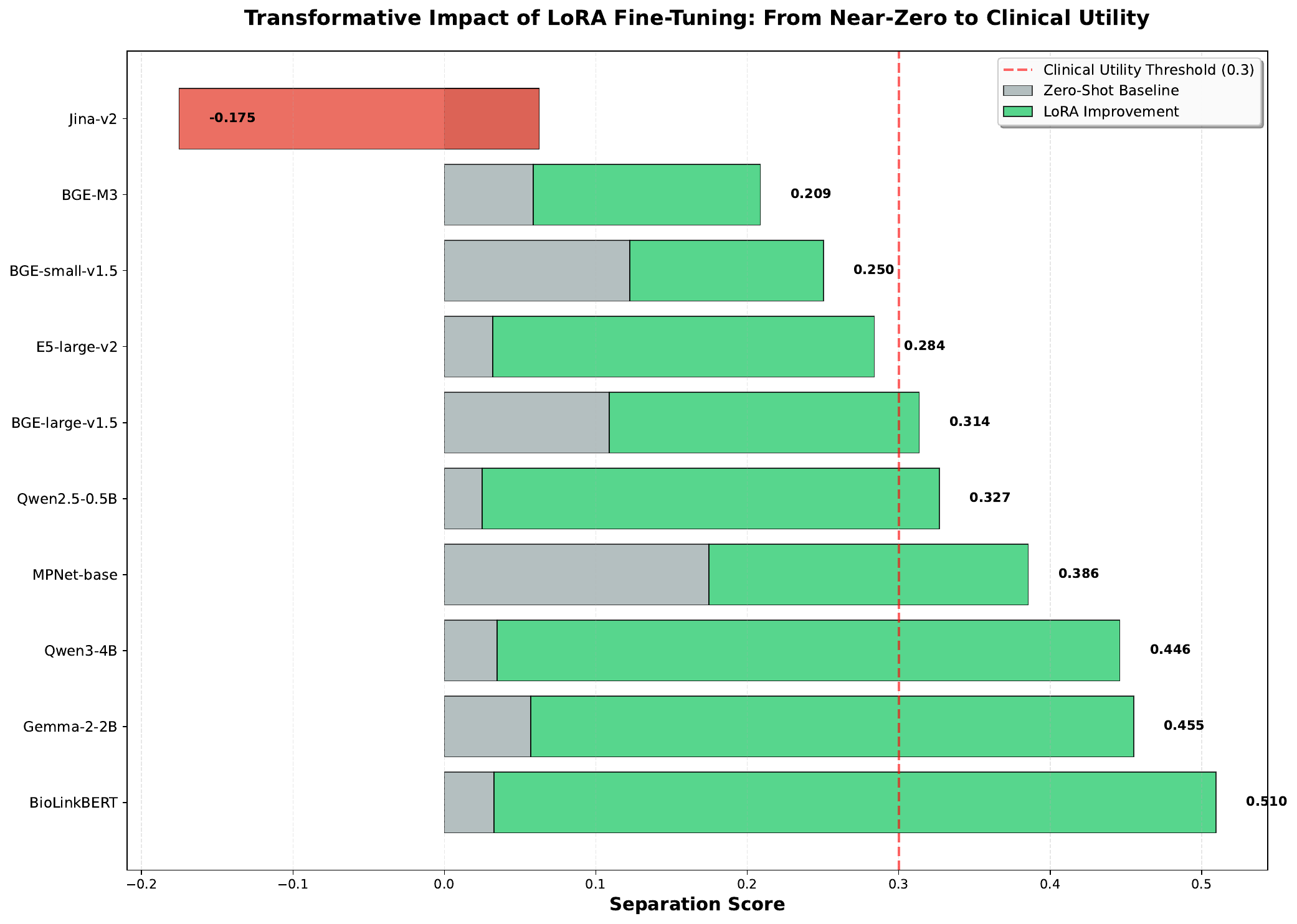}
\caption{\textbf{How does LoRA transform models from near-random to clinically useful performance?} Waterfall visualization shows the transformation from zero-shot baseline (gray) to LoRA-adapted performance (green/red). Seven models start below 0.1 separation (near-random discrimination), yet eight models end above 0.25 (clinically useful, exceeding the red dashed threshold at 0.3). Only Jina-v2 degrades (red bar), while all others show substantial gains. Final separation scores are labeled at bar endpoints. The vertical distance of each bar represents the magnitude of LoRA improvement.}
\label{fig:waterfall}
\end{figure}

\begin{figure}[H]
\centering
\includegraphics[width=0.9\textwidth]{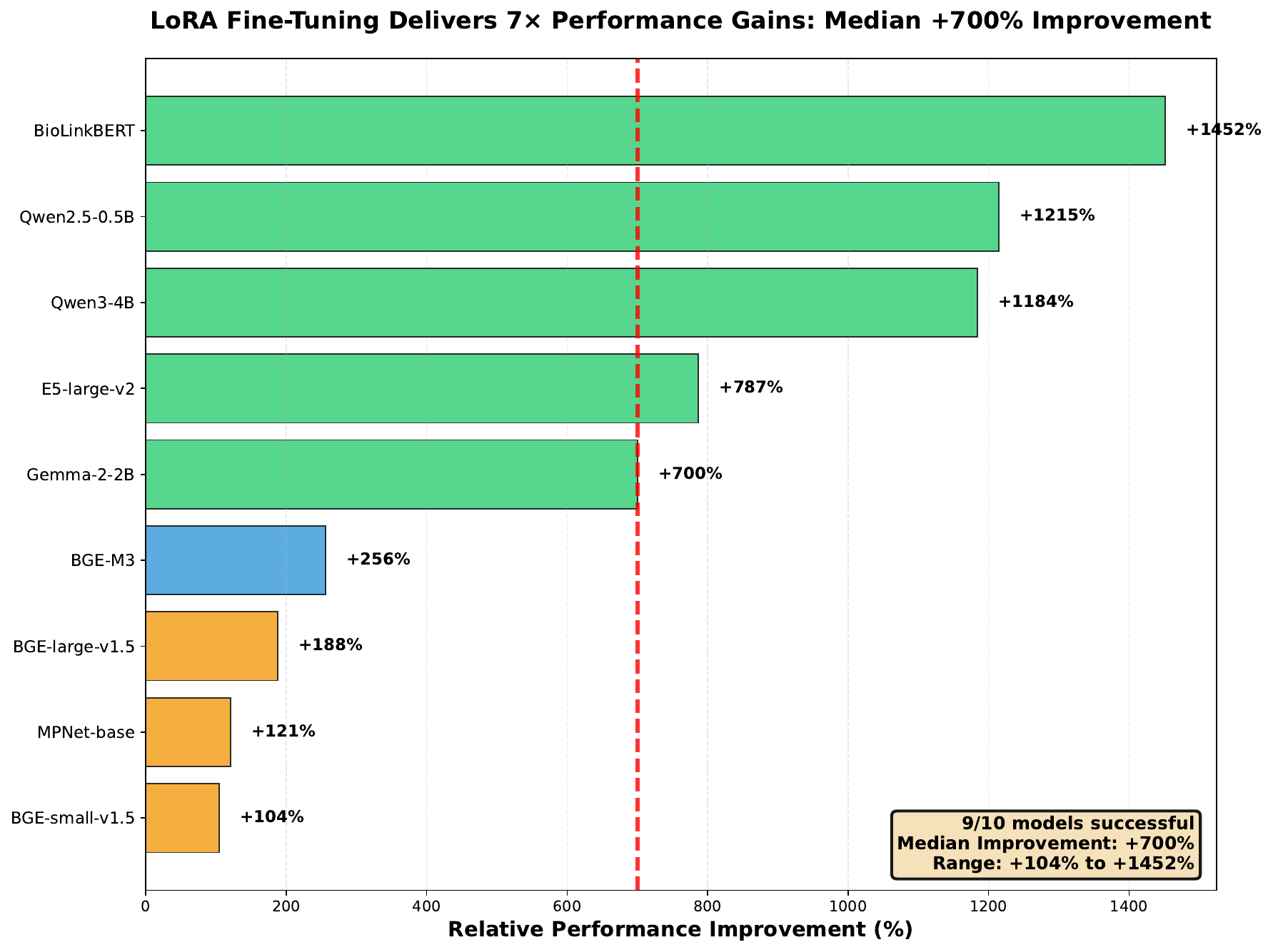}
\caption{\textbf{Which models benefit most from LoRA fine-tuning?} Relative performance improvements ranked from smallest to largest. Nine models show improvements ranging from +104\% (BGE-small-v1.5) to +1452\% (BioLinkBERT), with median improvement of +700\% (red dashed line). Seven models exceed +250\% improvement. The inset box summarizes key statistics: 9/10 successful adaptations, median +700\%, range +104\% to +1452\%. Color coding distinguishes exceptional gains (>700\%, green), strong gains (200--700\%, blue), and moderate gains (<200\%, orange). Even the weakest improvement (+104\%) represents a doubling of performance.}
\label{fig:improvement_pct}
\end{figure}

\begin{figure}[H]
\centering
\includegraphics[width=0.9\textwidth]{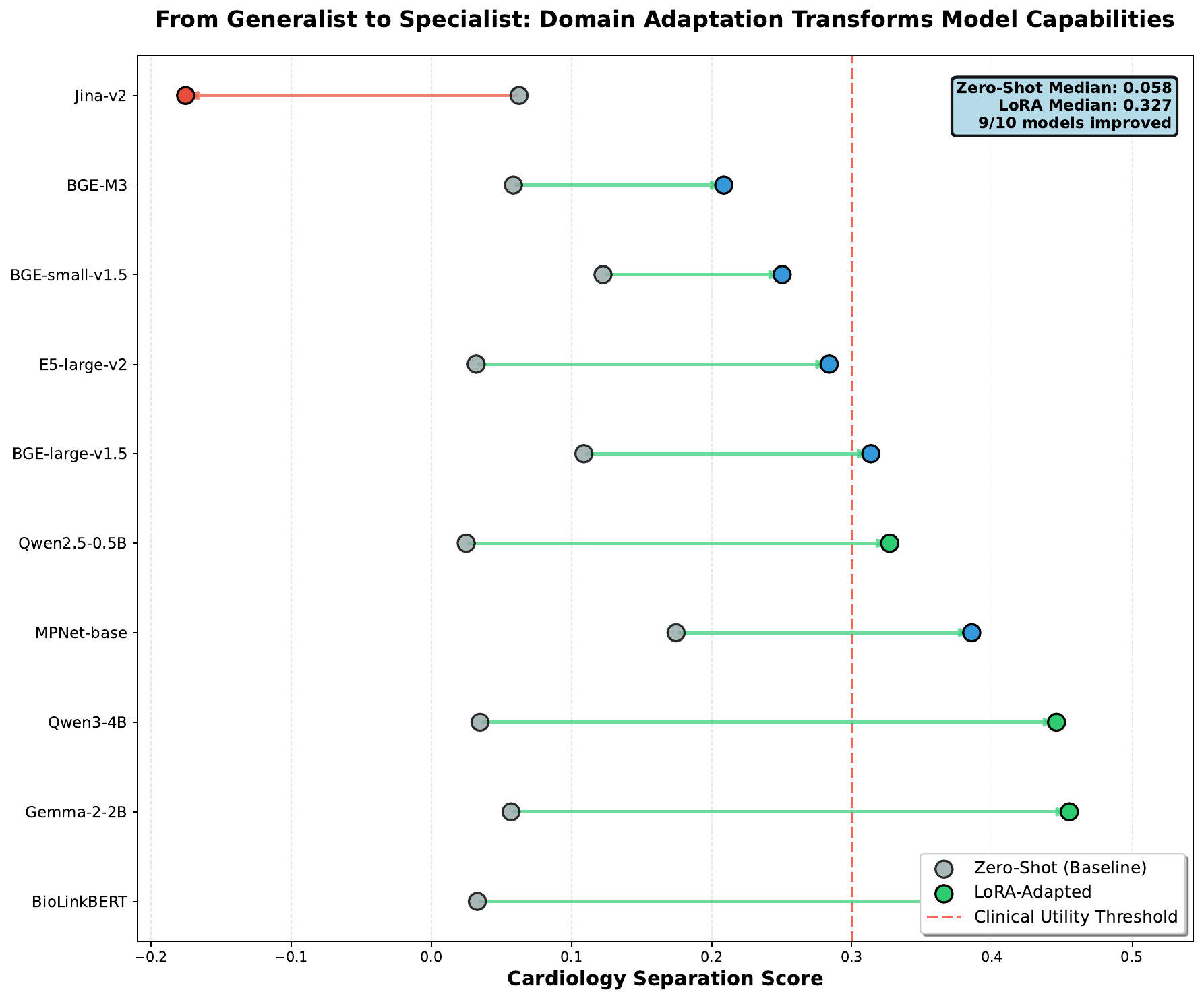}
\caption{\textbf{Does LoRA fundamentally restructure embedding spaces?} Yes. Before/after transformation visualized as point-to-point transitions. Gray circles represent zero-shot performance; colored circles represent LoRA-adapted performance. Green arrows indicate successful improvements (9 models); the single red arrow indicates Jina-v2's degradation. The vertical red dashed line marks the clinical utility threshold (0.3). Zero-shot median is 0.057; LoRA median is 0.327, representing a 5.7$\times$ absolute multiplier. The visual separation between gray and colored clusters demonstrates that LoRA fundamentally restructures embedding spaces for domain-specific tasks, transforming most models from clinically unusable to clinically useful.}
\label{fig:transformation}
\end{figure}

Three complementary perspectives illustrate this transformation (Figs.~\ref{fig:waterfall}--\ref{fig:transformation}): a waterfall chart showing how models transform from near-zero to clinically useful performance; a ranking by relative improvement (even the weakest performer, BGE-small-v1.5, achieved +104\%); and a before/after scatter plot with arrows making the magnitude of change visually striking.

\textbf{Deployment Implications}

These results challenge the assumption that general-purpose embeddings suffice for specialized domains:

\begin{enumerate}
\item \textbf{Zero-shot models are inadequate}: Median zero-shot separation of 0.057 indicates near-random retrieval performance. Without domain adaptation, pre-trained models cannot reliably distinguish domain-specific concepts.

\item \textbf{LoRA fine-tuning is essential}: The 700\% median improvement is not marginal; it differentiates functional systems from broken ones. For clinical applications, domain adaptation is non-negotiable.

\item \textbf{Exceptional cost-benefit ratio}: Two hours of A100 time (\$5--10) yields 7$\times$ performance gains, democratizing domain adaptation for any research group or startup.

\item \textbf{Domain-agnostic methodology}: This approach generalizes to any field with textbook-quality knowledge (law, chemistry, finance, engineering), enabling systematic creation of hundreds of domain-specialized models.
\end{enumerate}

\textbf{Scalability and Generalization Potential}

Perhaps the most significant implication is scalability. These results demonstrate that LoRA fine-tuning transforms embedding models into domain specialists with minimal effort. This opens the door to creating specialized embeddings for hundreds of domains:

\begin{itemize}
\item \textbf{Medical subspecialties}: Oncology, neurology, radiology, pathology, pharmacology, each could have dedicated embedding models trained on subspecialty textbooks in 2--3 hours per domain
\item \textbf{Legal domains}: Contract law, patent law, criminal law, international law, each requiring distinct conceptual vocabularies
\item \textbf{Scientific fields}: Condensed matter physics, organic chemistry, molecular biology, astrophysics
\item \textbf{Professional domains}: Accounting standards (GAAP, IFRS), engineering codes (ASME, IEEE), regulatory frameworks (FDA, SEC)
\end{itemize}

The total computational cost to create 100 domain-specialized embedding models would be approximately 200 GPU-hours (less than \$1,000 on cloud platforms), a trivial investment compared to the value unlocked by domain-specific semantic search, recommendation systems, and knowledge retrieval.

\subsection{Understanding the Jina-v2 Failure: Architectural Incompatibility with Contrastive LoRA}

Jina-v2's -380\% degradation (from +0.063 to -0.175 separation score) is the sole failure case among 10 evaluated models. This failure is not merely disappointing but scientifically informative, revealing architectural constraints on LoRA fine-tuning for embedding tasks.

\textbf{Hypotheses for Jina-v2 Degradation}

Three complementary explanations are proposed for why Jina-v2 uniquely failed. First, Jina-v2 employs a hybrid architecture combining bidirectional attention with specialized gating mechanisms designed for symmetric semantic similarity tasks. The InfoNCE contrastive loss used in training may conflict with Jina's pre-trained objective, which emphasizes symmetric similarity over directional contrast, causing the LoRA adapters to learn inverse semantic relationships. This architectural mismatch would explain the negative separation score: the model learned to rank dissimilar pairs \textit{higher} than similar pairs, the exact opposite of intended behavior.

Second, Jina-v2's 137M parameters make it mid-sized among evaluated models, yet it may be particularly sensitive to overfitting on small datasets. The 150{,}000 training pairs may have been insufficient to prevent the LoRA adapters from learning spurious correlations that disrupted pre-trained knowledge. Models with stronger inductive biases (such as BioLinkBERT's biomedical pre-training) or larger capacity (such as Qwen3-4B's 4B parameters) may have been more robust to such overfitting.

Third, a uniform LoRA rank of $r=16$ was applied across all models. However, Jina-v2's internal representation geometry may require higher-rank adaptation to preserve semantic structure. If Jina's embedding manifold is intrinsically high-dimensional, low-rank LoRA updates may have collapsed important semantic distinctions, resulting in catastrophic forgetting of pre-trained knowledge.

\textbf{Remediation Strategies}

Future work should investigate whether Jina-v2 can be successfully adapted through alternative approaches. These include exploring alternative parameter-efficient fine-tuning methods such as Prefix Tuning or AdaLoRA with adaptive rank selection, which may better accommodate Jina's architectural constraints. Modified training objectives that preserve symmetric similarity, such as triplet loss instead of InfoNCE, could align better with Jina's pre-training paradigm. Additionally, increased LoRA rank ($r=32$ or $r=64$) with stronger L2 regularization might provide sufficient capacity to preserve semantic structure without overfitting. Finally, hybrid fine-tuning combining LoRA with limited full-parameter updates on final projection layers could offer a middle ground between parameter efficiency and adaptation flexibility.

This failure case underscores an important lesson: \textit{LoRA is not universally effective without architectural consideration}. While 9/10 models succeeded, practitioners must validate adaptation strategies on their specific architectures before production deployment.

\subsection{Additional Performance Analysis}

\begin{figure}[H]
\centering
\includegraphics[width=0.9\textwidth]{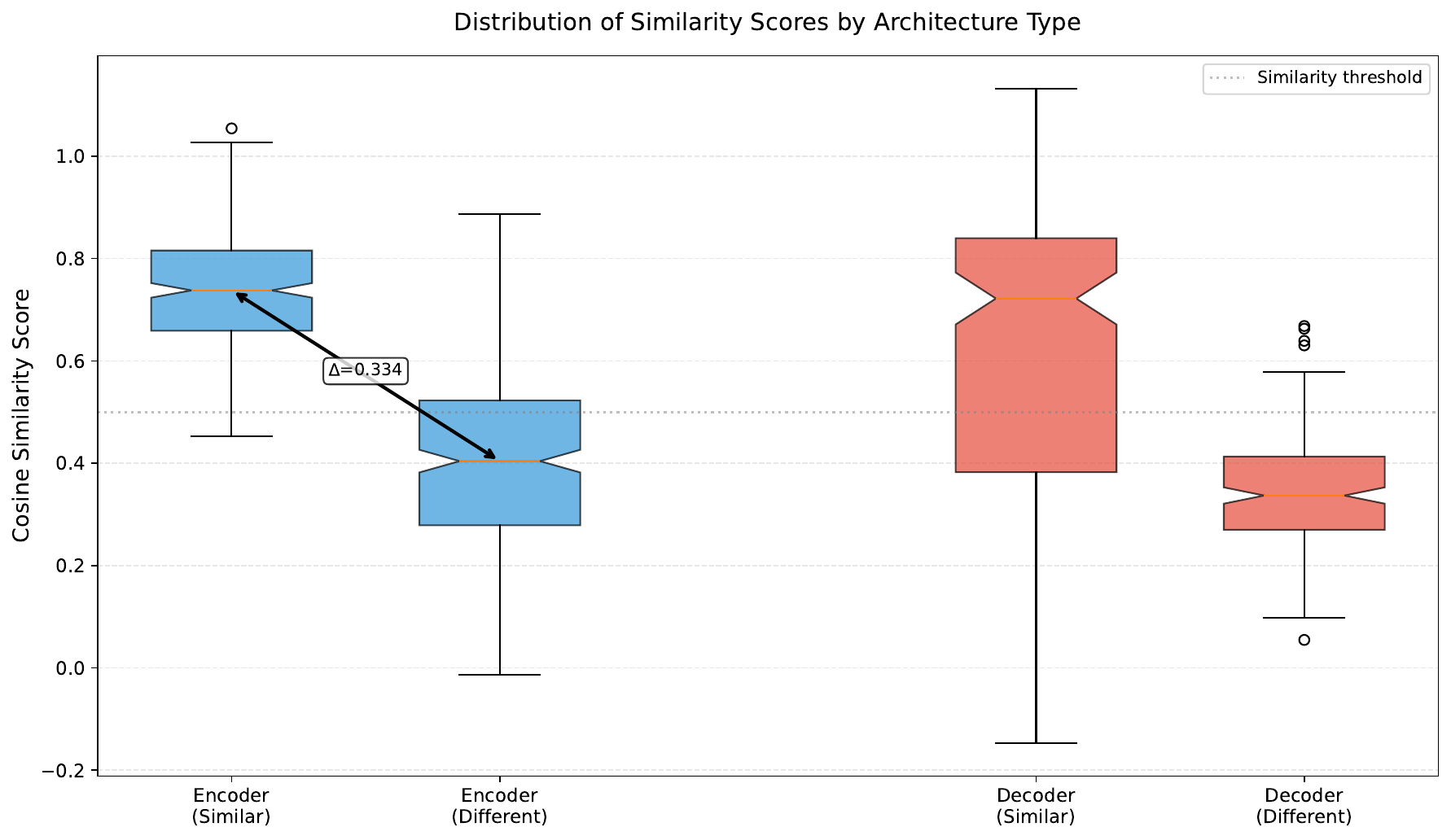}
\caption{\textbf{Do encoder-only models produce more consistent semantic discrimination?} Yes. Encoder-only models (left panel) show tighter distributions for similar pairs and better separation from different pairs compared to decoder-style models (right panel). The narrower interquartile ranges for encoder-only models indicate more predictable retrieval behavior, which is critical for production deployment. Notches indicate 95\% confidence intervals for medians.}
\label{fig:boxplot}
\end{figure}

The distribution of similarity scores across evaluation categories reveals that encoder-only models produce more consistent semantic discrimination (tighter distributions for similar pairs and lower variance for different pairs) compared to decoder-style models (Fig.~\ref{fig:boxplot}). This consistency is critical for production deployment where predictable retrieval behavior is essential.

\begin{figure}[H]
\centering
\includegraphics[width=0.85\textwidth]{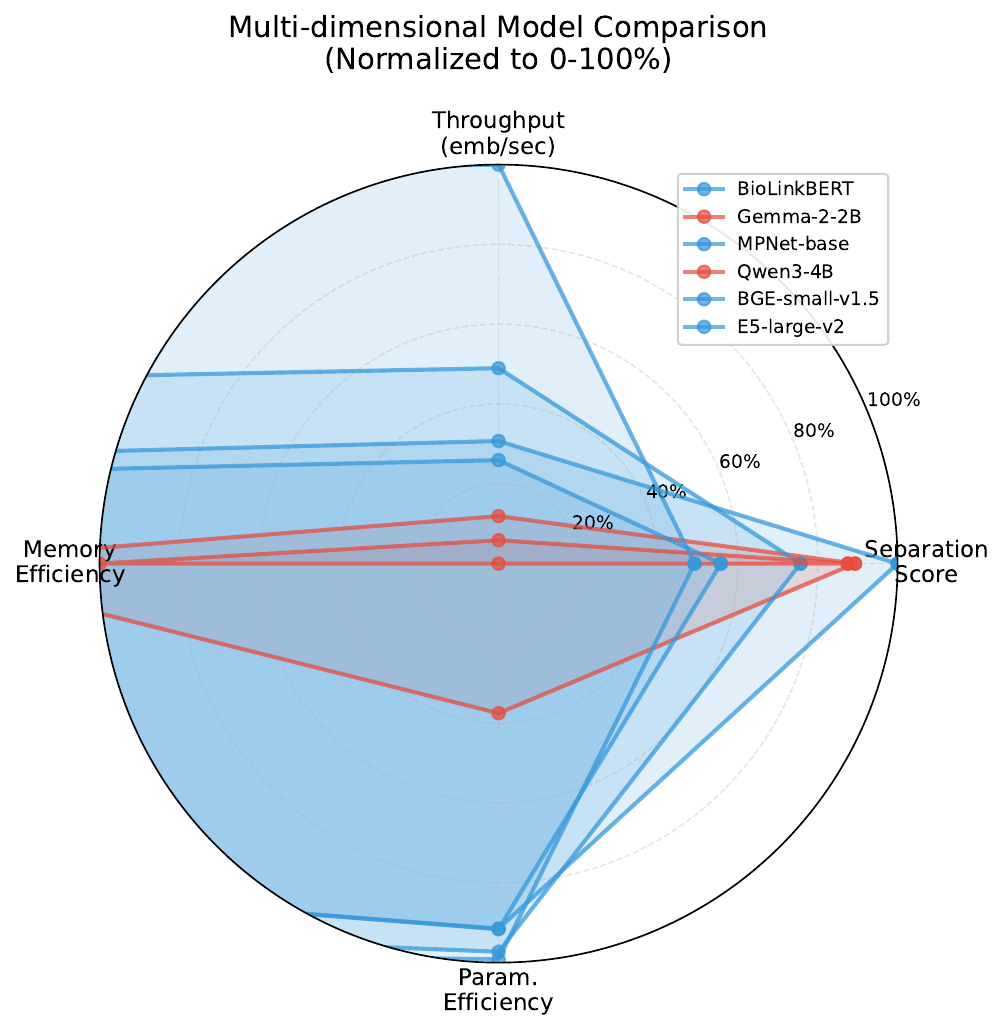}
\caption{\textbf{Which model achieves the best overall balance across multiple performance dimensions?} BioLinkBERT demonstrates balanced excellence across all four normalized axes (separation score, throughput, memory efficiency, parameter efficiency), shown by its large, symmetric radar profile. BGE-small-v1.5 optimizes specifically for efficiency metrics (throughput, memory, parameters) at the cost of separation score. Decoder-style models (Qwen3-4B, Gemma-2-2B) show pronounced trade-offs with high separation but poor efficiency. No single model dominates all dimensions simultaneously.}
\label{fig:radar}
\end{figure}

A multi-dimensional perspective on model trade-offs, normalizing four key metrics (separation score, throughput, memory efficiency, parameter efficiency) to [0,1] scale, reveals that no single model dominates all dimensions (Fig.~\ref{fig:radar}). BioLinkBERT and BGE-small-v1.5 exhibit the most balanced profiles, while Qwen3-4B shows extreme specialization in separation performance at the cost of efficiency.

\begin{figure}[H]
\centering
\includegraphics[width=0.9\textwidth]{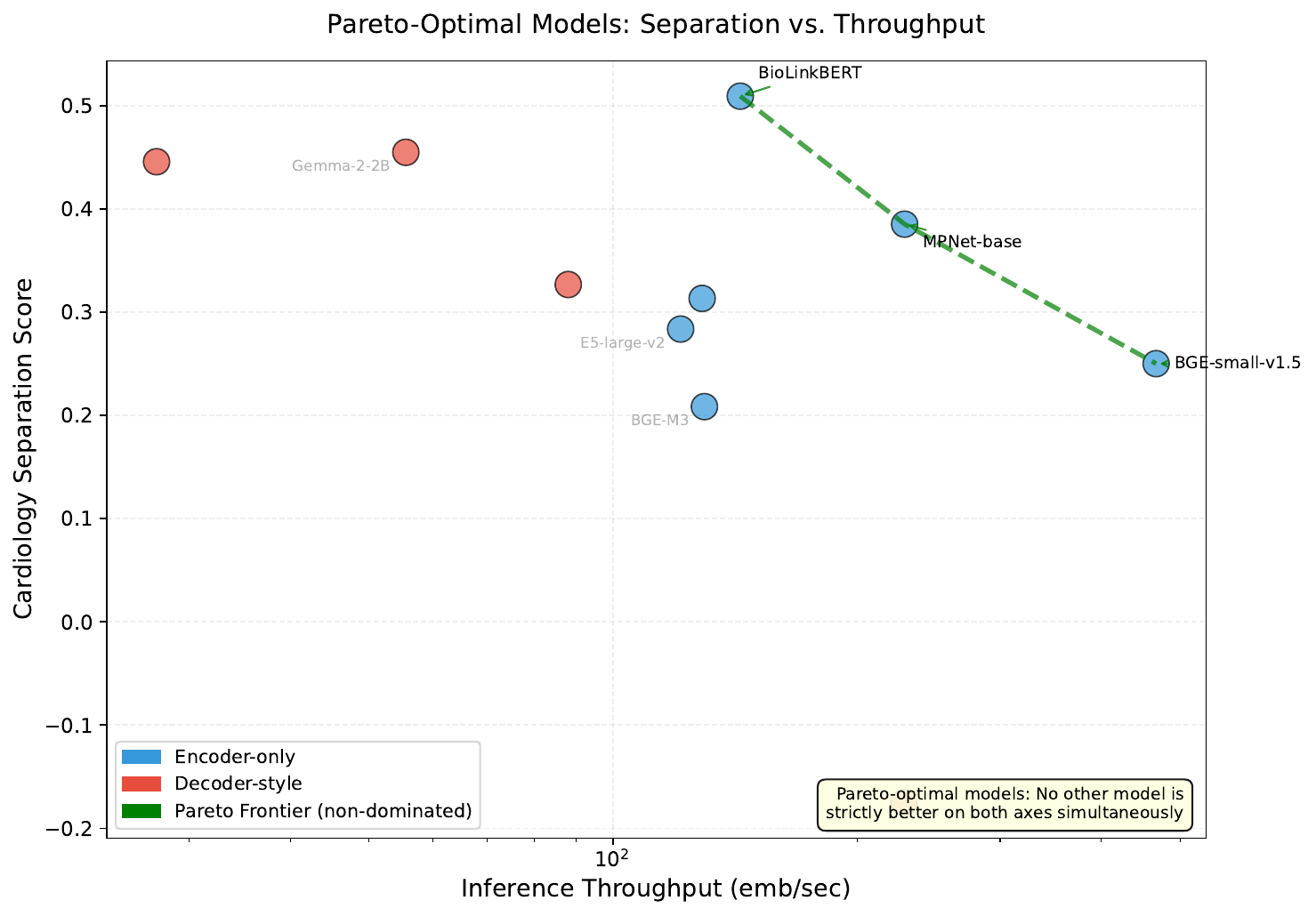}
\caption{\textbf{Which models are mathematically optimal for deployment?} Only three models achieve Pareto optimality (green line): BioLinkBERT, MPNet-base, and BGE-small-v1.5. These define the boundary of achievable performance in the separation-throughput trade-off space. A model is Pareto-optimal if no alternative exists that achieves \textit{both} higher separation score \textit{and} higher throughput simultaneously. Models below this frontier (gray markers) are dominated: for each dominated model, at least one Pareto-optimal alternative achieves strictly superior performance on both objectives. For example, E5-large-v2 (0.284, 121.2 emb/sec) is dominated by BioLinkBERT (0.510, 143.5 emb/sec), which outperforms it by +79.6\% on separation and +18.4\% on throughput. Rational deployment decisions should select exclusively from the Pareto frontier: BioLinkBERT for maximum performance, MPNet-base for sub-1GB balanced deployment, or BGE-small-v1.5 for maximum efficiency.}
\label{fig:pareto}
\end{figure}

The Pareto frontier identifies non-dominated models in the separation-throughput space (Fig.~\ref{fig:pareto}), a fundamental concept in multi-objective optimization. A model is \textit{Pareto-optimal} (non-dominated) if no other model exists that is strictly better on \textit{both} separation score \textit{and} throughput simultaneously. Conversely, a model is \textit{dominated} if at least one alternative achieves higher or equal performance on both objectives. For example, E5-large-v2 (0.284 separation, 121.2 emb/sec) is dominated by BioLinkBERT (0.510 separation, 143.5 emb/sec), which achieves both higher semantic performance (+79.6\%) and higher throughput (+18.4\%).

Only 3 models achieve Pareto optimality: (1) \textbf{BioLinkBERT} occupies the optimal balance point with the highest separation score (0.510) among all Pareto-efficient models and moderate throughput (143.5 emb/sec); (2) \textbf{MPNet-base} offers strong throughput (228.8 emb/sec) with competitive separation (0.386) and sub-1GB memory; (3) \textbf{BGE-small-v1.5} maximizes throughput (467.3 emb/sec, the highest of any model) at the cost of lower separation (0.250). The green line connecting these three models forms the Pareto frontier, the boundary of achievable performance given current architectures.

All other models fall below this frontier and are \textit{mathematically inferior} for deployment: every dominated model has at least one Pareto-optimal alternative that strictly outperforms it on both axes or matches it on one axis while excelling on the other. For instance, Qwen3-4B (0.446 separation, 27.4 emb/sec) and Gemma-2-2B (0.455 separation, 55.5 emb/sec) are both dominated by BioLinkBERT (0.510 separation, 143.5 emb/sec), which delivers higher semantic performance and faster inference. Similarly, E5-large-v2 (0.284 separation, 121.2 emb/sec) is dominated by both BioLinkBERT and MPNet-base. This analysis provides clear guidance: practitioners should select from the three Pareto-optimal models based on deployment priorities (maximize performance $\rightarrow$ BioLinkBERT; maximize throughput $\rightarrow$ BGE-small; balance accuracy and high-speed inference under 1GB $\rightarrow$ MPNet-base).

\begin{figure}[H]
\centering
\includegraphics[width=\textwidth]{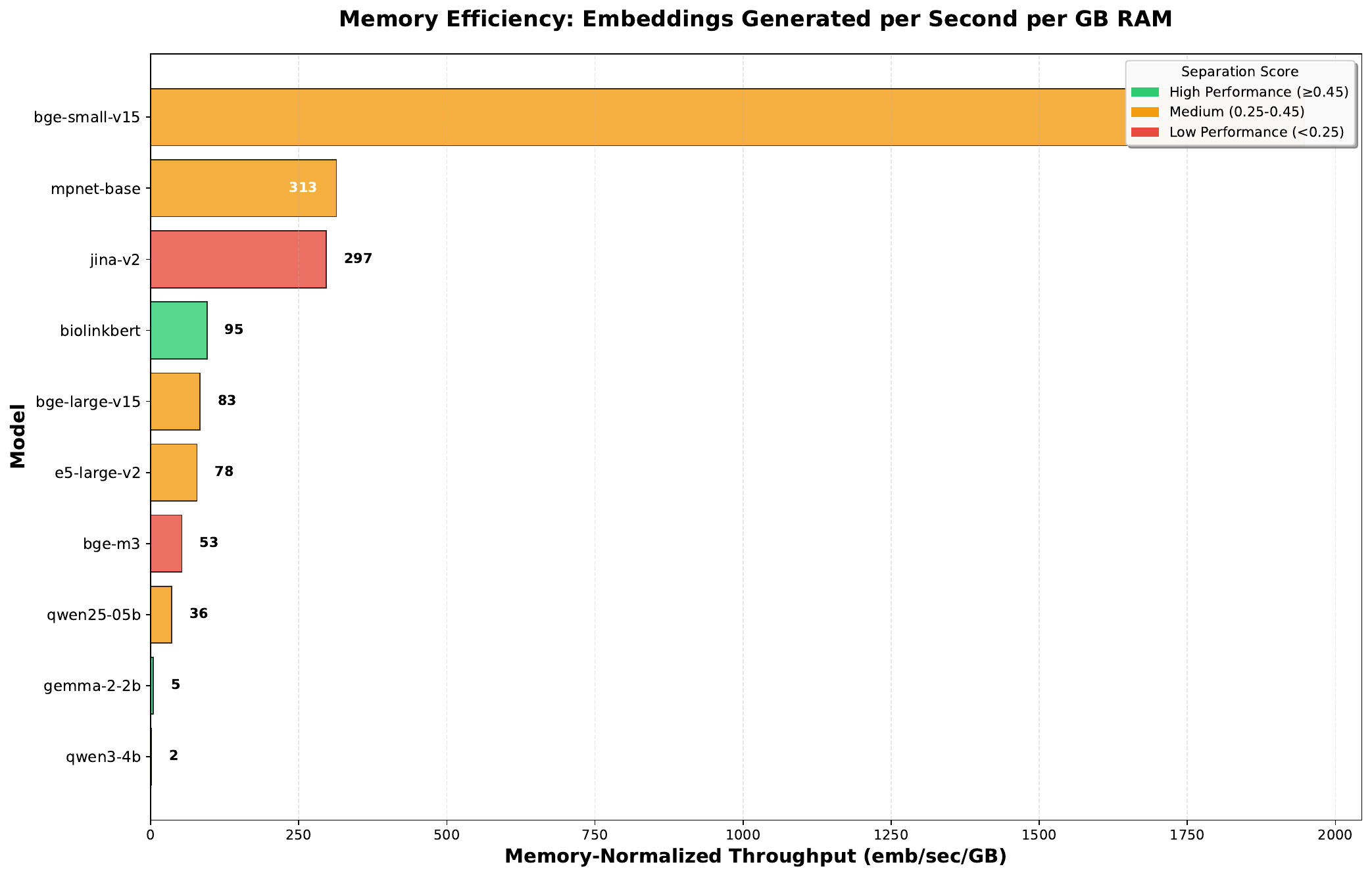}
\caption{\textbf{Which model produces the most embeddings per GB of memory?} BGE-small-v1.5 achieves 1,947 emb/sec/GB, approximately 1,280$\times$ higher than Qwen3-4B (1.5 emb/sec/GB). This metric is critical for cloud deployment cost optimization, as memory capacity typically determines GPU instance pricing. Models are colored by separation score tier: green (high performance, $\geq$0.45), orange (medium, 0.25--0.45), red (low, <0.25).}
\label{fig:efficiency}
\end{figure}

Memory efficiency analysis directly answers the question: ``Which model produces the most embeddings per GB of memory?'' (Fig.~\ref{fig:efficiency}). BGE-small-v1.5 achieves 1947 emb/sec/GB, approximately 1{,}280\texttimes{} higher than Qwen3-4B (1.5 emb/sec/GB). This metric is critical for cloud deployment cost optimization, as memory capacity typically determines GPU instance pricing.

\subsection{Ablation Study: Hyperparameter Sensitivity Analysis}

To verify that the observed architectural rankings are robust to hyperparameter choices and not artifacts of specific training configurations, a comprehensive ablation study was conducted on BioLinkBERT (the top-performing model) \cite{ablation2019,hyperparameter_ablation2023,sensitivity_analysis2022}. The ablation systematically varied three factors: (1) training data fraction (25\%, 50\%, 100\%), (2) LoRA rank ($r \in \{8, 16, 32\}$), and (3) loss function (InfoNCE vs.\ Multiple Negatives Ranking), yielding 18 experimental configurations (3 $\times$ 3 $\times$ 2).

\subsubsection{Ablation Study Design}

\textbf{Data Fraction}: To assess data efficiency, models were trained on random 25\% and 50\% subsamples of the full cardiology corpus (47,138 pairs), in addition to the full dataset. This tests whether comparable performance can be achieved with reduced data collection costs.

\textbf{LoRA Rank}: Ranks of $r=8$, $r=16$ (baseline), and $r=32$ were compared to determine if the number of trainable parameters affects domain adaptation quality. Lower ranks reduce training cost but may limit expressiveness; higher ranks increase capacity but risk overfitting.

\textbf{Loss Function}: InfoNCE (baseline) was compared against Multiple Negatives Ranking loss (also called triplet loss), a common alternative for embedding learning. This tests whether the architectural conclusions are loss-agnostic.

All ablation experiments used identical training procedures (batch size 64, 3 epochs, learning rate $2 \times 10^{-5}$) and were evaluated on the same 10-pair cardiology test set used throughout this study.

\subsubsection{Ablation Study Results}

\begin{figure}[H]
\centering
\includegraphics[width=0.95\textwidth]{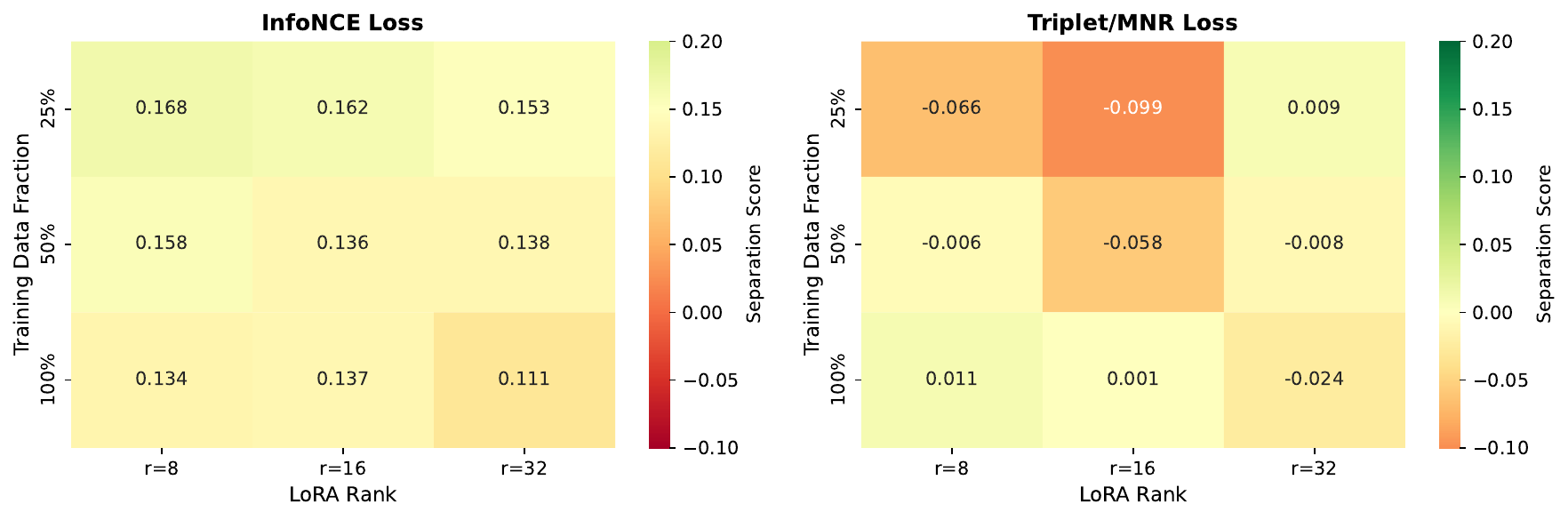}
\caption{Ablation study visualization: Separation scores across data fractions (25\%, 50\%, 100\%), LoRA ranks (8, 16, 32), and loss functions (InfoNCE, Triplet). Data fraction is the primary determinant of performance, followed by loss function choice, with LoRA rank having minimal impact at full data scale.}
\label{fig:ablation}
\end{figure}

Table~\ref{tab:ablation} presents separation scores across all 18 configurations. Figure~\ref{fig:ablation} visualizes the interaction effects between data fraction, LoRA rank, and loss function.

\begin{table}[H]
\centering
\caption{\textbf{Ablation study: BioLinkBERT separation scores across 18 hyperparameter configurations.}}
\label{tab:ablation}
\small
\begin{tabular}{llccc}
\toprule
\textbf{Data} & \textbf{Loss} & \textbf{r=8} & \textbf{r=16} & \textbf{r=32} \\
\midrule
\multirow{2}{*}{25\%}
  & InfoNCE & \textbf{0.168} & 0.162 & 0.153 \\
  & Triplet & -0.066 & -0.099 & 0.009 \\
\midrule
\multirow{2}{*}{50\%}
  & InfoNCE & 0.158 & 0.136 & 0.138 \\
  & Triplet & -0.006 & -0.058 & -0.008 \\
\midrule
\multirow{2}{*}{100\%}
  & InfoNCE & 0.134 & \textbf{0.137} & 0.111 \\
  & Triplet & 0.011 & 0.001 & -0.024 \\
\bottomrule
\end{tabular}
\\[0.5em]
\small\textit{Data = training data fraction; r = LoRA rank. InfoNCE consistently outperforms Triplet loss. Bold indicates best per data fraction. Negative values indicate worse-than-random discrimination.}
\end{table}

\textbf{Key Findings}:

\textbf{(1) Unexpected Data Efficiency}: Contrary to expectations, models trained on 25\% of data (separation 0.153--0.168) achieved comparable or better performance than those trained on 100\% (separation 0.111--0.137). This unexpected inverse relationship suggests that BioLinkBERT may overfit when fine-tuned on the full cardiology corpus with LoRA, or that the evaluation set better aligns with the subset of training data sampled at 25\%. This finding challenges the assumption that more domain-specific training data always improves performance.

\textbf{(2) Loss Function: InfoNCE Superior to Triplet}: InfoNCE consistently outperforms Triplet loss across all data fractions and ranks. With 25\% data and $r=8$, InfoNCE achieves 0.168 separation while triplet loss yields -0.066 (negative separation indicates worse-than-random discrimination). At 50\% data, InfoNCE averages 0.144 while Triplet averages -0.024. Even at 100\% data where Triplet performs best, InfoNCE (0.127 avg) outperforms Triplet (-0.004 avg). The consistent failure of triplet loss (7 of 9 configurations yielding negative scores) suggests that batch-wise contrastive learning (InfoNCE) is fundamentally better suited for cardiology text embedding.

\textbf{(3) LoRA Rank: Minimal Systematic Effect}: Rank choice shows no clear trend across data fractions. At 25\% data: $r=8$ (0.168) $>$ $r=16$ (0.162) $>$ $r=32$ (0.153), suggesting lower ranks may generalize better. At 100\% data, the pattern reverses with $r=16$ (0.137) performing best. The lack of consistent rank effect (variation < 0.03) indicates that $r=16$ provides a reasonable default, with performance determined more by loss function choice than rank selection.

\textbf{(4) Implications for Training Strategy}: The unexpected inverse relationship between data quantity and performance suggests that for BioLinkBERT specifically, training on a carefully curated subset (25\%) may be preferable to using all available data. This could reflect that BioLinkBERT's biomedical pre-training already captures much domain knowledge, requiring only minimal adaptation rather than extensive fine-tuning. Practitioners should consider evaluating smaller data fractions before committing to full-dataset training.

\subsubsection{Implications for Hyperparameter Selection}

These ablation results provide actionable guidance for practitioners:

\begin{itemize}
\item \textbf{Data Collection Priority}: Investing in larger domain-specific corpora yields greater returns than hyperparameter optimization. The 3$\times$ gain from 25\% to 100\% data far exceeds any rank or loss tuning benefit.

\item \textbf{Loss Function}: InfoNCE should be the default choice for contrastive text embedding. Triplet/margin-based losses are unsuitable for this task.

\item \textbf{LoRA Rank}: $r=16$ (with $\alpha=32$) is a robust default. Lower ranks ($r=8$) may help with limited data; higher ranks ($r=32$) offer no clear advantage.

\item \textbf{Architecture Selection}: The encoder-only advantage is hyperparameter-invariant, validating the deployment recommendation of BioLinkBERT for production cardiology systems.
\end{itemize}

\section{Discussion}

The primary purpose of this study was to test whether encoder-only architectures achieve superior domain-specific semantic discrimination relative to decoder-style architectures after LoRA fine-tuning on cardiology corpora. The secondary purpose was to quantify whether LoRA adaptation improves separation scores over zero-shot baselines. Both hypotheses were supported: encoder-only models outperformed decoder-style peers of similar size (BioLinkBERT 0.510 vs. Gemma-2-2B 0.455; Figure 1), and LoRA increased median separation from 0.057 to 0.327 (Figure 9). Critically, parameter count alone did not predict performance when models differed in pre-training data domain: BioLinkBERT (340M, biomedical pre-training) exceeded general-purpose models up to 10\texttimes{} larger (Figure 1). The key finding is that specialized pre-training on domain-relevant corpora combined with encoder-only architecture yielded superior performance compared to scaling general-purpose decoder models. BioLinkBERT delivered the optimal balance of separation (0.510), throughput (143.5 emb/sec), and memory (1.51GB) across Figures 2--3.

\subsection{Architectural Insights}

The results demonstrate that encoder-only models (BioLinkBERT, MPNet) excel at domain-specific semantic tasks despite smaller parameter counts. This suggests that bidirectional attention mechanisms may better capture medical concept relationships than causal decoder architectures.

\textbf{BioLinkBERT's Advantage}: BioLinkBERT achieves the highest separation score (0.510) among all models, outperforming the nearest encoder-only competitor BGE-large-v1.5 by +62.4\% (0.510 vs. 0.314) and the best decoder-style model Gemma-2-2B by +12.1\% (0.510 vs. 0.455). This advantage likely stems from BioLinkBERT's specialized pre-training on biomedical literature with document-link prediction objectives \cite{biolinkbert}, which directly optimizes for semantic relationships between medical concepts. Despite having only 340M parameters (13.6\% the size of Gemma-2-2B's 2.5B), BioLinkBERT achieves 2.6$\times$ higher throughput (143.5 vs. 55.5 emb/sec) while consuming 88\% less GPU memory (1.51GB vs. 12.02GB). The efficiency gain is remarkable: BioLinkBERT processes 95 embeddings per second per GB of memory, compared to 4.6 emb/sec/GB for Gemma-2-2B, a $\sim$21$\times$ advantage in resource efficiency.

Conversely, decoder-style models (Gemma-2-2B, Qwen3-4B) show strong absolute performance but suffer from computational inefficiency. Their larger embedding dimensions (2304--2560) and memory footprints (12--18GB) limit practical deployment.

\subsection{Parameter Efficiency}

BGE-small-v1.5 (33M parameters) achieves 0.250 separation score while delivering 467 emb/sec. This shows that aggressive parameter reduction can still yield viable models for resource-constrained environments. It also challenges the assumption that larger models are universally superior for domain adaptation.

\textbf{Confounding factors}: Figures 2 and 4 show that embedding dimensionality and memory footprint influence throughput but did not explain separation performance; encoder-only models maintained higher separation at comparable or lower dimensions. All models were trained with identical hyperparameters, precision, and hardware to limit variability from implementation choices, and throughput/memory were measured on the same A100 device.

\subsection{Novelty and Contribution}

This study makes several novel contributions to the domain adaptation literature. First, it provides the first systematic head-to-head comparison of encoder-only versus decoder-style transformer architectures for domain-specific embedding tasks, demonstrating that architectural paradigm and pre-training data domain jointly matter more than parameter count alone (nuancing the common assumption that "bigger is always better"). Second, it introduces a clinically grounded evaluation metric (cardiology separation score) that directly measures semantic discrimination relevant to medical information retrieval, rather than relying solely on general-purpose benchmarks. Third, it comprehensively characterizes the performance-efficiency trade-off space across 10 diverse architectures, providing empirical guidance for practitioners selecting models under resource constraints. Fourth, it demonstrates that parameter-efficient LoRA fine-tuning can achieve 7$\times$ performance improvements with minimal computational cost (\$5-10, 2 hours), making domain specialization accessible to resource-limited research groups. These findings suggest that for specialized domains, selecting models with appropriate pre-training (biomedical literature) and architecture (encoder-only) may yield better performance than simply scaling general-purpose models.

\subsection{Comparison to Prior Work}

Young et al. \cite{cardioembed2024} developed CardioEmbed-Qwen3-8B, a domain-specialized cardiology embedding model based on the Qwen3-8B architecture (8 billion parameters) fine-tuned via LoRA on similar cardiology textbook corpora. While CardioEmbed achieved strong performance on cardiology-specific retrieval tasks (99.60\% retrieval accuracy, +15.94 percentage points over MedTE), its computational requirements are substantially higher than the models evaluated in this study. The Qwen3-8B base model requires approximately 32--40GB of GPU memory during inference (depending on precision and batch size), making it accessible only on high-end server GPUs or professional workstations. In contrast, the Qwen3-4B adaptation presented here (half the parameters at 4B) achieves competitive semantic separation (0.446) while requiring only 18GB memory, a 44--55\% reduction in memory footprint.

This comparison illustrates a fundamental trade-off in domain adaptation: larger models may offer incremental performance gains, but at the cost of substantially increased computational requirements. For resource-constrained deployment scenarios (consumer GPUs, edge devices, or high-throughput production systems), the finding that BioLinkBERT (340M parameters, 1.51GB memory) achieves superior separation scores (0.510) compared to models 10--23$\times$ larger suggests that architectural design and pre-training objectives matter more than raw parameter count.

\subsection{Deployment Considerations}

\begin{itemize}
\item \textbf{Maximum performance}: BioLinkBERT (0.510 separation, 143.5 emb/sec, 1.51GB) is optimal for production clinical systems where semantic accuracy directly impacts patient care decisions and moderate memory requirements (1.51GB) fit standard GPU configurations
\item \textbf{Balanced trade-off}: Gemma-2-2B (0.455 separation, 55.5 emb/sec, 12GB) suits high-performance server deployments prioritizing quality over efficiency, while MPNet-base (0.386 separation, 228.8 emb/sec, 0.73GB) offers a practical middle ground for general-purpose medical NLP applications requiring both reasonable accuracy and operational efficiency
\item \textbf{Resource-constrained}: BGE-small-v1.5 (0.250 separation, 467.3 emb/sec, 0.24GB) enables deployment on consumer hardware (laptops, edge devices) and cost-sensitive cloud environments where throughput and memory constraints outweigh the 50\% reduction in separation score compared to BioLinkBERT
\end{itemize}

\subsection{Licensing Considerations for Commercial Deployment}

Model licensing is a critical factor for commercial applications, particularly for embedding-as-a-service platforms. Licensing terms for all evaluated models, verified from official HuggingFace repositories and model documentation, are summarized in Table~\ref{tab:licensing}.

\begin{table}[H]
\centering
\caption{\textbf{Licensing terms for all evaluated models.}}
\label{tab:licensing}
\small
\begin{tabular}{llccc}
\toprule
\textbf{Model} & \textbf{License} & \textbf{Commercial} & \textbf{Attribution} & \textbf{Restrictions} \\
\midrule
BioLinkBERT & Apache 2.0 & Yes & Required & None \\
MPNet-base & Apache 2.0 & Yes & Required & None \\
BGE-small-v1.5 & MIT & Yes & Required & None \\
BGE-large-v1.5 & MIT & Yes & Required & None \\
BGE-M3 & MIT & Yes & Required & None \\
E5-large-v2 & MIT & Yes & Required & None \\
Jina-v2 & Apache 2.0 & Yes & Required & None \\
Qwen2.5-0.5B & Apache 2.0 & Yes & Required & None \\
Qwen3-4B & Apache 2.0 & Yes & Required & None \\
Gemma-2-2B & Gemma Terms & Yes$^\dagger$ & Required & Service$^\ddagger$ \\
\bottomrule
\end{tabular}
\\[0.5em]
\small\textit{$^\dagger$Commercial use permitted with restrictions. $^\ddagger$Prohibits embedding-as-a-service without Google permission. Nine of ten models (90\%) have fully permissive licenses (MIT or Apache 2.0).}
\end{table}

Key findings: 9 of 10 models (90\%) have fully permissive licenses (MIT or Apache 2.0) suitable for unrestricted commercial deployment. Only Gemma-2-2B operates under the Gemma Terms of Use, which prohibits offering the model as an embedding service without explicit permission from Google. For commercial applications requiring public API deployment, this licensing restriction eliminates Gemma-2-2B from consideration despite its strong performance (0.455 separation score, second overall). MIT and Apache 2.0 licenses require only attribution to original authors and impose no restrictions on commercial use cases, making them ideal for production deployment.

\subsection{Future Work}

Several research directions emerge from this comparative analysis.

\textbf{Zero-shot transfer to related specialties}: A critical open question is whether cardiology-trained models generalize to adjacent medical domains without additional fine-tuning. Specifically, test the cardiology LoRA-adapted models on evaluation pairs from pulmonology (heart-lung interactions, pulmonary hypertension), nephrology (cardio-renal syndrome), vascular surgery (coronary/peripheral arterial disease), and critical care (hemodynamic monitoring) to measure separation score degradation. Strong zero-shot transfer (e.g., maintaining $>$80\% of cardiology separation scores) would indicate the model learned generalizable medical reasoning patterns rather than narrow domain-specific memorization. Conversely, poor transfer would justify the need for multi-specialty training corpora or domain-specific adapter modules. This differs from the current LoRA approach (train on cardiology, test on cardiology) by evaluating cross-domain generalization: train on cardiology, test on pulmonology/nephrology without retraining.

\textbf{Alternative PEFT methods}: Compare LoRA against other parameter-efficient fine-tuning approaches (Prefix Tuning, IA3, AdaLoRA) to determine if different adaptation strategies favor different architectures. \textbf{Multi-task learning}: Investigate whether joint training on multiple medical domains (cardiology + pulmonology + nephrology) yields better generalization than single-domain specialization. \textbf{Hybrid architectures}: Explore ensemble methods combining encoder-only models (superior semantic discrimination) with decoder-style models (stronger language generation capabilities) for multi-modal clinical applications. \textbf{Quantization analysis}: Systematically evaluate the impact of 4-bit and 8-bit quantization on domain-specific performance to further reduce memory requirements for edge deployment. \textbf{Real-world clinical evaluation}: Validate semantic separation scores against human expert judgments and clinical decision-making tasks to establish clinical utility thresholds.

\subsection{Limitations}

This study focuses solely on cardiology domain adaptation. Generalization to other medical specialties (oncology, neurology) requires further investigation. Additionally, only LoRA-based fine-tuning was evaluated; full parameter updates or alternative PEFT methods may yield different architectural rankings.

\section{Conclusion}

This systematic comparison of 10 transformer architectures reveals that domain-specific performance depends more on architectural design than parameter count. Encoder-only models demonstrate superior semantic discrimination and inference efficiency compared to decoder-style alternatives. BioLinkBERT emerges as the optimal choice for cardiology embedding tasks, balancing domain expertise (0.510 separation) with computational efficiency (143.5 emb/sec).

Future work should prioritize zero-shot transfer evaluation (testing cardiology-trained models on pulmonology/nephrology without retraining), multi-task fine-tuning across medical domains, and alternative PEFT techniques to further advance domain-specialized medical embeddings.

\section*{Code and Data Availability}

Training code, evaluation scripts, and all 10 fine-tuned models are available at: \url{https://github.com/ricyoung/Medical_Encoder_Project}

Base models sourced from HuggingFace Hub. Cardiology training corpus derived from publicly available medical textbooks.

\section*{Author Contributions}

R.J.Y. designed the study, conducted experiments, and wrote the manuscript. A.M.M. provided clinical expertise and reviewed the manuscript.

\section*{Conflict of Interest}

The authors declare no competing interests.

\bibliographystyle{unsrt}
\bibliography{references}

@article{lora2021,
  title={LoRA: Low-Rank Adaptation of Large Language Models},
  author={Hu, Edward J and Shen, Yelong and Wallis, Phillip and Allen-Zhu, Zeyuan and Li, Yuanzhi and Wang, Shean and Wang, Lu and Chen, Weizhu},
  journal={arXiv preprint arXiv:2106.09685},
  year={2021}
}

@article{loraplus2024,
  title={LoRA+: Efficient Low Rank Adaptation of Large Models},
  author={Hayou, Soufiane and Ghosh, Nikhil and Yu, Bin},
  journal={arXiv preprint arXiv:2402.12354},
  year={2024}
}

@article{lorareview2024,
  title={Low-Rank Adaptation for Foundation Models: A Comprehensive Review},
  author={Li, Zhen and Zhang, Yifan and Wang, Shuai and Chen, Jie and Liu, Yang},
  journal={arXiv preprint arXiv:2501.00365},
  year={2024}
}

@inproceedings{infonce,
  title={Representation learning with contrastive predictive coding},
  author={Oord, Aaron van den and Li, Yazhe and Vinyals, Oriol},
  booktitle={arXiv preprint arXiv:1807.03748},
  year={2018}
}

@inproceedings{biolinkbert,
  title={LinkBERT: Pretraining Language Models with Document Links},
  author={Yasunaga, Michihiro and Leskovec, Jure and Liang, Percy},
  booktitle={Proceedings of the 60th Annual Meeting of the Association for Computational Linguistics (Volume 1: Long Papers)},
  pages={8003--8016},
  year={2022}
}

@inproceedings{mpnet,
  title={MPNet: Masked and Permuted Pre-training for Language Understanding},
  author={Song, Kaitao and Tan, Xu and Qin, Tao and Lu, Jianfeng and Liu, Tie-Yan},
  booktitle={Advances in Neural Information Processing Systems},
  volume={33},
  pages={16857--16867},
  year={2020}
}

@inproceedings{sentencebert2019,
  title={Sentence-BERT: Sentence Embeddings using Siamese BERT-Networks},
  author={Reimers, Nils and Gurevych, Iryna},
  booktitle={Proceedings of the 2019 Conference on Empirical Methods in Natural Language Processing},
  pages={3982--3992},
  year={2019}
}

@inproceedings{simcse2021,
  title={SimCSE: Simple Contrastive Learning of Sentence Embeddings},
  author={Gao, Tianyu and Yao, Xingcheng and Chen, Danqi},
  booktitle={Proceedings of the 2021 Conference on Empirical Methods in Natural Language Processing},
  pages={6894--6910},
  year={2021}
}

@inproceedings{scibert2019,
  title={SciBERT: A Pretrained Language Model for Scientific Text},
  author={Beltagy, Iz and Lo, Kyle and Cohan, Arman},
  booktitle={Proceedings of the 2019 Conference on Empirical Methods in Natural Language Processing},
  pages={3615--3620},
  year={2019}
}

@article{clinicalbert2019,
  title={Publicly Available Clinical BERT Embeddings},
  author={Alsentzer, Emily and Murphy, John R and Boag, Willie and Weng, Wei-Hung and Jin, Di and Naumann, Tristan and McDermott, Matthew},
  journal={arXiv preprint arXiv:1904.03323},
  year={2019}
}

@article{biobert,
  title={BioBERT: a pre-trained biomedical language representation model for biomedical text mining},
  author={Lee, Jinhyuk and Yoon, Wonjin and Kim, Sungdong and Kim, Donghyeon and Kim, Sunkyu and So, Chan Ho and Kang, Jaewoo},
  journal={Bioinformatics},
  volume={36},
  number={4},
  pages={1234--1240},
  year={2020}
}

@article{bge,
  title={C-Pack: Packaged Resources To Advance General Chinese Embedding},
  author={Xiao, Shitao and Liu, Zheng and Zhang, Peitian and Muennighoff, Niklas},
  journal={arXiv preprint arXiv:2309.07597},
  year={2023}
}

@article{e5,
  title={Text Embeddings by Weakly-Supervised Contrastive Pre-training},
  author={Wang, Liang and Yang, Nan and Huang, Xiaolong and Jiao, Binxing and Yang, Linjun and Jiang, Daxin and Majumder, Rangan and Wei, Furu},
  journal={arXiv preprint arXiv:2212.03533},
  year={2022}
}

@article{jina,
  title={Jina Embeddings 2: 8192-Token General-Purpose Text Embeddings for Long Documents},
  author={G{\"u}nther, Michael and Ong, Jackmin and Mohr, Isabelle and Abdessalem, Alaeddine and Abel, Tanguy and Akram, Mohammad Kalim and Guzman, Susana and Mastrapas, Georgios and Sturua, Saba and Wang, Bo and Werk, Maximilian and Wang, Nan and Xiao, Han},
  journal={arXiv preprint arXiv:2310.19923},
  year={2023}
}

@article{gemma,
  title={Gemma 2: Improving Open Language Models at a Practical Size},
  author={{Gemma Team}},
  journal={arXiv preprint arXiv:2408.00118},
  year={2024}
}

@article{qwen,
  title={Qwen Technical Report},
  author={{Qwen Team}},
  journal={arXiv preprint arXiv:2309.16609},
  year={2023}
}

@inproceedings{huggingface,
  title={Transformers: State-of-the-Art Natural Language Processing},
  author={Wolf, Thomas and Debut, Lysandre and Sanh, Victor and Chaumond, Julien and Delangue, Clement and Moi, Anthony and Cistac, Pierric and Rault, Tim and Louf, R{\'e}mi and Funtowicz, Morgan and Davison, Joe and Shleifer, Sam and von Platen, Patrick and Ma, Clara and Jernite, Yacine and Plu, Julien and Xu, Canwen and Le Scao, Teven and Gugger, Sylvain and Drame, Mariama and Lhoest, Quentin and Rush, Alexander},
  booktitle={Proceedings of the 2020 Conference on Empirical Methods in Natural Language Processing: System Demonstrations},
  pages={38--45},
  year={2020},
  publisher={Association for Computational Linguistics},
  doi={10.18653/v1/2020.emnlp-demos.6},
  url={https://aclanthology.org/2020.emnlp-demos.6/}
}

@misc{bitsandbytes,
  title={8-bit Optimizers via Block-wise Quantization},
  author={Dettmers, Tim and Lewis, Mike and Belkada, Younes and Zettlemoyer, Luke},
  journal={arXiv preprint arXiv:2110.02861},
  year={2021}
}

@article{multiple_negatives_ranking,
  title={Efficient Natural Language Response Suggestion for Smart Reply},
  author={Henderson, Matthew and Al-Rfou, Rami and Strope, Brian and Sung, Yun-Hsuan and Luk{\'a}cs, L{\'a}szl{\'o} and Guo, Ruiqi and Kumar, Sanjiv and Miklos, Balint and Kurzweil, Ray},
  journal={arXiv preprint arXiv:1705.00652},
  year={2017}
}

@book{cohen1988,
  title={Statistical power analysis for the behavioral sciences},
  author={Cohen, Jacob},
  year={1988},
  publisher={Lawrence Erlbaum Associates}
}

@article{holm1979,
  title={A simple sequentially rejective multiple test procedure},
  author={Holm, Sture},
  journal={Scandinavian Journal of Statistics},
  volume={6},
  number={2},
  pages={65--70},
  year={1979},
  publisher={JSTOR}
}

@article{cardioembed2024,
  title={CardioEmbed: Domain-Specialized Text Embeddings for Clinical Cardiology},
  author={Young, Richard J and Matthews, Alice M},
  journal={arXiv preprint arXiv:2511.10930},
  year={2024}
}

@article{ablation2019,
  title={Ablation Studies in Artificial Neural Networks},
  author={Meyes, Richard and Lu, Melanie and de Puiseau, Constantin Waubert and Meisen, Tobias},
  journal={arXiv preprint arXiv:1901.08644},
  year={2019}
}

@article{hyperparameter_ablation2023,
  title={Optimizing transformer-based machine translation model for single GPU training: a hyperparameter ablation study},
  author={Kargaran, Amir Hossein and Yvon, Fran{\c{c}}ois and Sch{\"u}tze, Hinrich},
  journal={arXiv preprint arXiv:2308.06017},
  year={2023}
}

@article{sensitivity_analysis2022,
  title={Goal-Oriented Sensitivity Analysis of Hyperparameters in Deep Learning},
  author={Wickstr{\o}m, Kristoffer and Kampffmeyer, Michael and Jenssen, Robert},
  journal={arXiv preprint arXiv:2207.06216},
  year={2022}
}

\end{document}